%% file: main.tex
\documentclass{article}

\usepackage{arxiv}
\usepackage[utf8]{inputenc} 
\usepackage[T1]{fontenc}  
\usepackage[acronym]{glossaries}
\usepackage{hyperref}       
\usepackage{url}            
\usepackage{booktabs}       
\usepackage{amsfonts}       
\usepackage{nicefrac}       
\usepackage{microtype}      
\usepackage{lipsum}		
\usepackage{adjustbox}
\usepackage{graphicx}
\usepackage{natbib}
\usepackage{doi}
\usepackage{listings}
\usepackage{subcaption}
\usepackage{xcolor}
\usepackage{graphicx}
\usepackage{multirow}
\usepackage{fancyhdr}
\usepackage{float}
\lstdefinestyle{pytorch}{
    language=Python,
    backgroundcolor=\color{white},
    commentstyle=\color{gray},
    keywordstyle=\color{blue},
    stringstyle=\color{red},
    basicstyle=\ttfamily,
    breaklines=true,
    frame=single,
}
\usepackage{authblk} 
\date{}
\title{OASST-ETC Dataset: Alignment Signals from Eye-tracking Analysis of LLM Responses \thanks{ACM 2025. This is the author's version of the work. It is posted here for your personal use. Not for redistribution. The definitive version was published in PACMHCI, \url{https://doi.org/10.1145/3725840}.}}

\author[1,2]{Angela Lopez-Cardona}
\author[1]{Sebastian Idesis}
\author[1]{Miguel Barreda-Ángeles}
\author[2]{Sergi Abadal}
\author[1]{Ioannis Arapakis}

\affil[1]{Telefónica Scientific Research, Barcelona, Spain}
\affil[2]{Universitat Politècnica de Catalunya, Barcelona, Spain}

\affil{\texttt{angela.lopezcardona@telefonica.com, sebastianariel.idesis@telefonica.com, ioannis.arapakis@telefonica.com}}

\input{glossary}
\begin{document}

\maketitle
\begin{abstract}
While Large Language Models (LLMs) have significantly advanced natural language processing, aligning them with human preferences remains an open challenge. Although current alignment methods rely primarily on explicit feedback, eye-tracking (ET) data offers insights into real-time cognitive processing during reading. In this paper, we present OASST-ETC, a novel eye-tracking corpus capturing reading patterns from 24 participants, while evaluating LLM-generated responses from the OASST1 dataset. Our analysis reveals distinct reading patterns between preferred and non-preferred responses, which we compare with synthetic eye-tracking data. Furthermore, we examine the correlation between human reading measures and attention patterns from various transformer-based models, discovering stronger correlations in preferred responses. This work introduces a unique resource for studying human cognitive processing in LLM evaluation and suggests promising directions for incorporating eye-tracking data into alignment methods. The dataset and analysis code are publicly available.
\end{abstract}

\input{01_introduction}
\input{02_background}
\input{03_related_work}

\input{04_corpus}
\input{05_analyse}

\input{06_analyse_attention}

\input{07_conclusions}

\section*{Ethical statement}

We obtained informed consents from the participants, ensuring the anonymization of their data. Consequently, the eye-tracking datasets collected in this study provide anonymous records in compliance with ethical board approvals, containing no personal information. This approach guarantees the privacy and anonymity of the collected data. To address potential privacy concerns associated with eye-tracking data collection and processing, the datasets presented are unlinkable to individual participants, thus preventing any possible misuse.

\subsubsection*{Acknowledgments}
This research is supported by Horizon Europe's European Innovation Council through the Pathfinder program (SYMBIOTIK project, grant 101071147) and by the Industrial Doctorate Plan of the Department of Research and Universities of the Generalitat de Catalunya, under Grant AGAUR 2023 DI060.

\bibliographystyle{unsrtnat}
\bibliography{bibliography}

\newpage
\input{appendix}

\end{document}

%% file: glossary.tex
\makeglossaries
\newacronym{ai}{AI}{Artificial Intelligence}
\newacronym{aoi}{AOI}{Area of Interest}
\newacronym{bilstm}{BiLSTM}{Bidirectional Long Short-Term Memory}
\newacronym{cot}{CoT}{Chain of Thought}
\newacronym{ct}{CT}{Conversation Tree}
\newacronym{cnn}{CNN}{Convolutional Neural Network}
\newacronym{cv}{CV}{Computer Vision}
\newacronym{drl}{DRL}{Deep Reinforcement Learning}
\newacronym{ddpg}{DDPG}{Deep Deterministic Policy Gradient}
\newacronym{dl}{DL}{Deep Learning}
\newacronym{dnn}{DNN}{Deep Neural Networks}
\newacronym{dqn}{DQN}{Deep Q-learning}
\newacronym{dpo}{DPO}{Direct Preference Optimization}
\newacronym{ddqn}{DDQN}{Double Q-learning}
\newacronym{et}{ET}{Eye-tracking}
\newacronym{ft}{FT}{Fine-Tuning}
\newacronym{gae}{GAE}{Generalized Advantage Estimate}
\newacronym{gnn}{GNN}{Graph Neural Networks}
\newacronym{gqa}{GQA}{Grouped Query Attention}
\newacronym{hrlaif}{HRLAIF}{Hybrid Reinforcement Learning from AI Feedback}
\newacronym{il}{IL}{Imitation Learning}
\newacronym{ipopt}{IPOPT}{Interior Point Optimizer}
\newacronym{kl}{KL}{Kullback–Leibler}
\newacronym{llm}{LLM}{Large Language Model}
\newacronym{llms}{LLMs}{Large Language Models}
\newacronym{lstm}{LSTM}{Long short-term memory}
\newacronym{lora}{LoRA}{Low-Rank Adaptation}
\newacronym{lr}{LR}{Learning Rate}
\newacronym{mdp}{MDP}{Markov Decision Process}
\newacronym{mha}{MHA}{multi-head attention}
\newacronym{ml}{ML}{Machine Learning}
\newacronym{mlp}{MLP}{Multilayer Perceptron}
\newacronym{mpnn}{MPNN}{Message Passing Neural Networks}
\newacronym{mtl}{MTL}{Multi-task learning}
\newacronym{nlp}{NLP}{Natural Language Processing}
\newacronym{nn}{NN}{Neural Networks}
\newacronym{ocr}{OCR}{Optical Character Recognition}
\newacronym{orms}{ORMs}{Outcome-supervised reward models}
\newacronym{p3o}{P3O}{Pairwise Proximal Policy Optimization}
\newacronym{peft}{PEFT}{Parameter-Efficient Fine-Tuning}
\newacronym{pi}{PI}{Policy Iteration}
\newacronym{ppo}{PPO}{Proximal Policy Optimization}
\newacronym{prms}{PRMs}{Process Based Reward Models}
\newacronym{raft}{RAFT}{Reward rAnked FineTuning}
\newacronym{rag}{RAG}{Retrieval-Augmented Generation}
\newacronym{rnn}{RNN}{Recurrent Neural Network}
\newacronym{rnns}{RNNs}{Recurrent Neural Networks}
\newacronym{rl}{RL}{Reinforcement Learning}
\newacronym{rlaif}{RLAIF}{Reinforcement Learning from AI Feedback}
\newacronym{rlcd}{RLCD}{Reinforcement Learning from Contrastive Distillation}
\newacronym{rlhf}{RLHF}{Reinforcement Learning from Human Feedback}
\newacronym{rm}{RM}{Reward Model}
\newacronym{rms}{RMs}{Reward Models}
\newacronym{rso}{RSO}{Statistical Rejection Sampling Optimization}
\newacronym{rrhf}{RRHF}{Rank Responses to align Human Feedback}

\newacronym{sota}{SOTA}{State of the Art}
\newacronym{sft}{SFT}{Supervised Fine-Tuning}
\newacronym{smoe}{SMoE}{Sparse Mixture of Experts}
\newacronym{slic}{SLiC}{Sequence Likelihood Calibration}
\newacronym{slcihf}{SLiC-HF}{Sequence Likelihood Calibration with Human Feedback}
\newacronym{sorms}{SORMs}{Stepwise ORMs}
\newacronym{swa}{SWA}{sliding window attention}
\newacronym{trpo}{TRPO}{Trust Region Policy Optimization}
\newacronym{trt}{TRT}{total reading time}

\newacronym{vi}{VI}{Value Iteration}

%% file: 01_introduction.tex

\section{Introduction}


\acrfull{llms} have advanced the field of \acrfull{nlp} and opened new possibilities across a range of tasks \cite{touvron_llama_2023, bai_constitutional_2022, ouyang_training_2024, dubey_llama_2024}. However, to date, a prevalent challenge remains in aligning \acrshort{ai} systems, and \acrshort{llms} in particular, with human values, intentions, and preferences. A popular approach to addressing this challenge leverages explicit human feedback to guide model preferences, with \acrfull{rlhf} introduced by OpenAI \cite{ouyang_training_2024} being the most widely adopted method. Specifically, \acrshort{rlhf} has been applied across state-of-the-art \acrshort{llms}, both in proprietary models such as OpenAI's GPT-4 \cite{openai_gpt-4_2023} and Anthropic's Claude \cite{bai_constitutional_2022}, as well as open-source models like Meta's Llama 3 \cite{meta_ai_llama3model_cardmd_2024}.

In this context, a significant problem is obtaining high-quality data to train \acrshort{llms} effectively \cite{casper_open_2023}, as well as designing a \acrfull{rm} that accurately ``embodies'' human values. The complexity of this task can lead to misaligned reward functions, steering models towards learning unintended or incomplete objectives, a problem known as ``reward hacking'' \cite{ji_ai_2024}. Optimal methods for gathering feedback to align \acrshort{llms} with human goals remain an open question \cite{casper_open_2023}. Traditionally, explicit feedback is collected after users have reviewed model outputs. However, human decision making is inherently complex and involves a comprehensive evaluation of diverse information types before taking action. As a result, users often navigate through various conscious and subconscious \cite{bargh1999, nisbett1977} processes and stages \cite{anand1988behavioural}, which can frequently lead to misalignment \cite{Malkoc2005}, thus relying on self-reported assessments that may not reflect the cognitive processes involved in real-time language understanding. By contrast, methods such as \acrfull{et} offer high temporal and spatial data resolution \cite{zhang_eye-tracking_2024}, and are more robust against cognitive biases which demonstrate their connection with explicit feedback, such as aesthetic evaluations or subjective preferences \cite{li_uniar_2024}.

\acrshort{et} data has been shown to add value in various \acrshort{nlp} tasks, as demonstrated by prior work \cite{huang_longer_2023, khurana_synthesizing_2023, hollenstein_advancing_2019, yang_plm-as_2023, kiegeland_pupil_2024, deng_pre-trained_2023, mathias_eyes_2018, mcguire_sentiment_2021}, and recent research has begun exploring its applications in human alignment for \acrshort{llms} \cite{kiegeland_pupil_2024, lopez-cardona_seeing_2024}. 
Despite its potential, this area remains largely unexplored. Acquiring eye-tracking data in labs is challenging due to the need for costly equipment and privacy concerns \citep{khurana_synthesizing_2023}. To address these issues, researchers are using generative models to produce synthetic eye-tracking data. The performance of these models greatly improves when training datasets align closely with specific NLP tasks \citep{maharaj_eyes_2023, sood_improving_2020, huang_longer_2023}, highlighting the need for task-specific eye-tracking datasets to enhance model accuracy and effectiveness in \acrshort{nlp}. Finally, studies have shown strong correlations between human eye movements and attention patterns in Transformer-based models \citep{wang_gaze-infused_2024, bensemann_eye_2022, sood_interpreting_2020}. In this setting, our work presents the following \textbf{research questions} and contributions:

\begin{itemize}

    \item \textbf{RQ1: How do human reading patterns differ between chosen and rejected responses within the LLMs human alignment framework?} We develop and release the OASST-ETC (OASST-\acrlong{et} Corpus)\footnote{\url{https://github.com/Telefonica-Scientific-Research/oasstetc}.}, a first-of-its-kind \acrshort{et} dataset that captures the eye movements of 24 participants as they read and evaluate responses generated with the OASST1 \cite{kopf_openassistant_2023} dataset. This unique corpus supports the exploration of how \acrshort{et} data can enhance alignment training and reveal different patterns in these responses.

    \item \textbf{RQ2: Are these differences in reading patterns also evident in synthetic eye-tracking data?} We perform a parallel analysis using synthetically generated reading measures to provide a comparative viewpoint/perspective.

    \item \textbf{RQ3: How do these reading measures correlate with model-generated attention patterns in Transformer-based models across different architecture types and model tasks?} For the first time in human alignment tasks, we compare human attention patterns with model-generated attention, analysing models of varying architectures and trained in different tasks, providing insights into how human preferences in \acrshort{llm} outputs align with model-generated attention mechanisms.
\end{itemize}

%% file: 02_background.tex
\section{Background}

\subsection{Eye tracking for \acrshort{nlp}}
\label{subsec:eye}

Eye-tracking technology has enabled the study of human visual attention and behaviour in a non-invasive and portable manner. This method enables precise tracking of eye movements in response to stimuli, proving valuable across fields such as psychology, marketing, medical diagnostics, and human-computer interaction. Reading, as a complex information processing activity, involves numerous cognitive processes \citep{bolliger_emtec_2024}, and analyzing these can offer insights into text comprehension \citep{mathias_survey_2020}.

While reading, our eyes perform two main actions: (1) fixations, when the eyes pause momentarily to process information, and (2) saccades, which are rapid shifts where visual information is not acquired \cite{hollenstein_zuco_2018}. Eye-tracking devices gather raw data about scanpaths, frequency and duration of fixations, as well as pupillary dilation variations. From this raw input, reading measures (i.e. eye-tracking features) are derived, like fixation duration, or aggregate measures like \acrfull{trt}. These metrics allow for a comprehensive and robust description of reading behaviour \cite{hollenstein_zuco_2018}. \autoref{tab:et-features} presents some key word-level reading measures, though the literature also covers several other measures.

\begin{table}[t!]
\renewcommand{\arraystretch}{0.6}
\caption{\acrfull{et} reading measures per word \citep{hollenstein_cmcl_2021}}
\centering
\begin{adjustbox}{width=0.9\textwidth}
\small
\begin{tabular}{lll}
\toprule
\textbf{Acronym} & \textbf{Measure} & \textbf{Definition} \\ \midrule 
 FFD & First Fixation Duration & Time spent on the initial fixation\\
 GPT & Go-Past Time &  Cumulative fixation time before moving to the right\\
 TRT & Total Reading Time & Overall time spent fixating on a word\\
 nFix & Number of Fixations & Number of fixations on each word\\ 
 fixProp & Proportion of participants & Proportion of participants that fixated on the word \\\bottomrule 
\end{tabular}
\end{adjustbox}
\label{tab:et-features}
\end{table}

One of the disadvantages of collecting organic eye-tracking data is the need for proprietary, high-precision equipment, along with data privacy concerns \cite{khurana_synthesizing_2023}. Moreover, gaze data is generally unavailable during inference. To overcome these limitations, researchers have developed generative models that predict gaze patterns for specific text stimuli, supporting various \acrshort{nlp} applications. Models trained on open-source datasets, such as Eyettention \cite{deng_eyettention_2023}, SCANDL \cite{bolliger_scandl_2023}, and ScanTextGAN \cite{khurana_synthesizing_2023}, simulate fixation sequences or scanpaths. Other models, such as those proposed by \citet{li_torontocl_2021, hollenstein_multilingual_2021} and the token-level tasks introduced by \citet{hollenstein_cmcl_2021, hollenstein_cmcl_2022}, directly predict reading metrics at the token level. These models are further applied in research; for instance, Eyettention \cite{deng_eyettention_2023} has been used in studies such as \cite{deng_fine-tuning_2024, deng_pre-trained_2023}, and the model from \cite{li_torontocl_2021} was applied in \cite{zhang_eye-tracking_2024, wang_gaze-infused_2024}. These predictive models can be applied directly or fine-tuned further. Fine-tuning typically involves using an eye-tracking dataset specific to the NLP task, incorporating cognitive signals or using the pretrained model as a foundation to optimize for a target task, ultimately infusing NLP objectives with useful inductive biases \cite{deng_fine-tuning_2024, deng_pre-trained_2023, maharaj_eyes_2023, sood_improving_2020}. Alternatively, modules can be trained to encode both textual and cognitive signals in parallel. Inference in such models requires only text input, enabling the learned relationships to be applied without needing cognitive input \citep{ren_cogalign_2021, ding_cogbert_2022}.


\subsection{Human alignment in \acrlong{llms}} 
\label{sec:backgroun_llms}

\acrshort{llms} are initially trained as foundation models on large, general-purpose datasets to capture broad language patterns. After this pretraining phase, they undergo further refinement through human alignment methods, which enhance their relevance, responsiveness, and ability to address user-specific needs. To date, \acrshort{rlhf} \citep{ouyang_training_2024} remains the primary alignment technique used in state-of-the-art \acrshort{llms} like GPT-4 \citep{openai_gpt-4_2023}, Claude \citep{bai_constitutional_2022}, Bard \citep{google_google_2023}, and Llama 2-Chat \citep{touvron_llama_2023}. Implementations of \acrshort{rlhf} vary in data gathering, training procedures, and choice of \acrshort{rl} algorithms, yet they generally involve three main steps \citep{ouyang_training_2024}: (1) collecting human feedback, (2) training a \acrlong{rm} on this feedback, and (3) fine-tuning the \acrshort{llms} using \acrshort{rl} methods, such as \acrfull{ppo} \citep{schulman_proximal_2017}, guided by the trained \acrshort{rm}. Additional modifications to \acrshort{rlhf} approaches include using more detailed rewards \citep{bai_constitutional_2022, wu_fine-grained_2023, dong_steerlm_2023, wang_helpsteer_2023, wang_helpsteer2_2024} or exploring alternative \acrshort{rl} algorithms \citep{wu_pairwise_2023}.

An alternative to \acrshort{rlhf} is \acrshort{dpo} \citep{rafailov_direct_2023}, which optimizes language models based on preference data without a dedicated \acrshort{rm}, leveraging an offline \acrshort{rl} approach. Another widely adopted technique is statistical rejection sampling, also referred to as best-of-N or top-k-over-N \citep{bai_constitutional_2022, touvron_llama_2023, dubey_llama_2024}. Additionally, some human alignment techniques discard \acrshort{rl} entirely to avoid instability issues, and instead fine-tune models on filtered samples selected by a \acrshort{rm} or other sources \citep{dong_raft_2023, yuan_rrhf_2023}.

Regardless of the specific approach, human alignment techniques typically rely on datasets comparing preferred versus rejected responses to the same prompt. Recent methods have introduced more complex reward structures, such as separately evaluating aspects like helpfulness and harmlessness \citep{bai_constitutional_2022}, or implementing finer-grained preference models \citep{wu_fine-grained_2023, dong_steerlm_2023, wang_helpsteer_2023}. These datasets generally include human preference labels that differentiate across preferred ($y_w$) and rejected ($y_l$) responses for a given prompt ($x$): $\mathcal{D} = \{(x^{(i)}, y_w^{(i)}, y_l^{(i)})\}_{i=1}^N$.

While approaches such as \acrshort{ppo} may differ in implementation, most methods leverage these datasets to train the \acrshort{rm} as a stand-in for direct human feedback. In the original \acrshort{rlhf} implementation \citep{ouyang_training_2024}, the \acrshort{rm} functions as a classifier, predicting the probability of a preferred response, $p^{*}$, between two responses (\autoref{eq:3}) using the Bradley-Terry model \citep{bradley_rank_1952}. By training copies of the original language model on labeled preference data, the \acrshort{rm} can gauge user preference between different text outputs \citep{lambert_rewardbench_2024}.

\begin{equation}
    \label{eq:3}
    p^{*}(y_w \succ y_l \mid x) = \frac{\exp(r^{*}(x, y_w))}{\exp(r^{*}(x, y_w)) + \exp(r^{*}(x, y_l))}.
\end{equation}

Ultimately, the effectiveness of language model alignment depends on the quality of the \acrlong{rm} \citep{pace_west--n_2024} and the preference datasets, as these are key to shaping a chatbot's behavior \citep{shen_trickle-down_2023}. A prevalent challenge across alignment techniques is securing high-quality data \citep{casper_open_2023}, including issues such as evaluator bias, supervision complexities, and feedback reliability \citep{casper_open_2023}. A promising solution is \acrfull{rlaif} \citep{bai_constitutional_2022}, where \acrshort{rms} are trained using preference data generated by other \acrshort{llms}, enabling scalable, cost-effective data collection. Recent research has explored various forms of RLAIF \citep{lee_rlaif_2023, jiao_starling-7b_2023, cui_ultrafeedback_2024, li_hrlaif_2024, yang_rlcd_2024}, expanding reward modeling's scope to include AI-generated feedback and emphasizing the importance of accurately capturing user preferences.

%% file: 03_related_work.tex
\section{Related work} 
\label{sec:rw}

\subsection{Eye-tracking corpus} 
\label{sec:rw_corpus}
Several widely-used datasets for eye-tracking research are publicly available, including ZUCO \cite{hollenstein_zuco_2020}, ZUCO2 \cite{hollenstein_zuco_2018}, PROVO \cite{luke_provo_2018}, ETSA-I \cite{mishra_predicting_2016}, ETSA-II \cite{mishra_cognition-cognizant_2018}, GECO \cite{cop_presenting_2017}, GECO-MT \cite{colman_geco-mt_2022}, CELER \cite{berzak_celer_2022}, and CopCo \cite{hollenstein_copenhagen_2022}, InteRead \cite{zermiani_interead_2024} and PoTeC \citep{jakobi_potec_2024}. Additionally, multilingual datasets such as MECO \cite{siegelman_expanding_2022} and EGGBD5 \cite{mathias_happy_2020} are becoming increasingly available. Recently developed datasets like WebQAmGaze \cite{ribeiro_webqamgaze_2023} include new methods, such as recording gaze data with a webcam, which could simplify data collection. However, no dataset has yet been specifically created for human alignment in \acrshort{llms}. Although English-language data is most common, multilingual resources are expanding, with most datasets focusing on naturalistic reading or specific tasks, such as sentiment analysis. For instance, the ITB-HGC:IITB dataset \cite{maharaj_eyes_2023} is designed to address hallucination detection, a significant challenge in \acrshort{llms}, and EMTEC \cite{bolliger_emtec_2024}, which utilizes text generated by \acrshort{llms} but focuses on general reading rather than the selection of responses.

\subsection{Eye-tracking applications for \acrshort{nlp}} 
\label{sec:rw_et_nlp}

\acrlong{et} has found applications across numerous \acrshort{nlp} tasks, from named entity recognition \citep{hollenstein_entity_2019, ren_cogalign_2021, yu_ceer_2024, hollenstein_advancing_2019} and text comprehension \citep{ahn_towards_2020, reich_inferring_2022, sood_improving_2020} to language modeling \citep{huang_longer_2023, huang_long-range_2023, deng_eyettention_2023} and question answering \citep{zhang_eye-tracking_2024, wang_gaze-infused_2024}. Other examples include code comprehension \citep{alakmeh_predicting_2024}, code summarization \citep{zhang_eyetrans_2024}, and hallucination detection \citep{maharaj_eyes_2023}. The use of \acrshort{et} in sentiment analysis and sarcasm detection tasks has also received a lot of attention, as suggested by the body of related work \citep{mishra_predicting_2016, mishra_leveraging_2016, mishra_cognition-cognizant_2018, barrett_sequence_2018, huang_longer_2023, khurana_synthesizing_2023, hollenstein_advancing_2019, yang_plm-as_2023, kiegeland_pupil_2024, deng_pre-trained_2023, mathias_eyes_2018, mcguire_sentiment_2021}. Recent research has also investigated connections between \acrshort{et} features and human alignment. For instance, \cite{kiegeland_pupil_2024} developed a dataset using \acrshort{et} signals for \acrshort{dpo}, building on sentiment generation frameworks from \cite{deng_pre-trained_2023} and \cite{yang_plm-as_2023}. Similarly, \cite{lopez-cardona_seeing_2024} demonstrated that incorporating \acrshort{et} signals can improve the accuracy of \acrshort{rm}.

\subsection{Human- vs. model-based relative attention comparison} \label{sec:rw_attention}

The relationship between human reading patterns and attention mechanisms in \acrshort{dl} models has been studied extensively to enhance both cognitive interpretability and explainability in language models \cite{deng_eyettention_2023, bolliger_emtec_2024}. Researchers have examined various architectures, including encoder models like BERT and RoBERTa \cite{hollenstein_relative_2021, wang_gaze-infused_2024}, as well as T5 \cite{eberle_transformer_2022} and GPT \cite{wu_eye_2024, wang_probing_2024}.

Specifically, studies have analysed attention across model layers, from early \cite{hollenstein_relative_2021} to final layers \cite{sood_interpreting_2020}, and across multiple layers \cite{wang_gaze-infused_2024}, using gaze reading measures like Total Reading Time (TRT) \cite{hollenstein_relative_2021, eberle_transformer_2022} and other metrics \cite{sood_interpreting_2020, wang_gaze-infused_2024, wang_probing_2024}. Methods for comparison include Spearman correlation \cite{wang_gaze-infused_2024, bensemann_eye_2022} and KL divergence \cite{sood_interpreting_2020}, revealing various patterns in human and model attention alignment. Alternative attention measures, such as flow attention \cite{eberle_transformer_2022} and gradient-based saliency \cite{hollenstein_relative_2021, wu_eye_2024}, have also shown promising results. Additionally, studies have examined how aligning a model's training task with the task used in eye-tracking corpora impacts attention patterns, often yielding varied findings \cite{wu_eye_2024, sood_interpreting_2020, eberle_transformer_2022}. Building upon these studies, our work provides the first systematic comparison between attention mechanisms and eye-tracking data for human alignment tasks.

%% file: 04_corpus.tex
\section{OASST-ETC dataset creation}

We collected eye-tracking data from 24 participants (\autoref{subsec:particpants}) while they read and evaluated pairs of responses to prompts generated by a \acrshort{llm}. The stimuli for this study were drawn from the OASST1 dataset \citep{kopf_openassistant_2023} (\autoref{subsec:stimuli}). Following established best practices \citep{jakobi_reporting_2024, dunn_minimal_2024}, we ensure open access to data across processing stages: (1) raw data (\autoref{subsec:data_adquisition}), (2) filtered fixation sequences (\autoref{subsec:eye_features}), and (3) computed reading measures (\autoref{sec:compute_readingm}). Additionally, electrodermal activity (EDA) data gathered during the experiment will also be included with the dataset release (a detailed description is shown in Appendix \ref{sec:app_eda}).

\subsection{Stimulus} 
\label{subsec:stimuli}

For this study, we used the OpenAssistant Conversations dataset (OASST1)\footnote{\url{https://huggingface.co/datasets/OpenAssistant/oasst1}} \citep{kopf_openassistant_2023}, a comprehensive, crowd-sourced, human-generated, and annotated conversation dataset widely applied in \acrshort{llms} human alignment research \citep{kopf_openassistant_2023, dettmers_qlora_2023, wu_meta-rewarding_2024}. OASST1 consists of 161,443 messages across 35 languages, with 461,292 quality ratings and over 10,000 annotated conversation trees. To construct our \acrshort{et} corpus, we filtered the dataset to include only English-language texts, selected prompt and response pairs suitable for our display parameters (\autoref{subsec:data_adquisition}), and identified the two most contrasting responses for each prompt. We selected 360 prompts and their corresponding 720 responses that met these criteria.

\subsection{Participants} 
\label{subsec:particpants}

Twenty-six individuals participated in our study (16 male, 8 female; mean age 31.66 $\pm$ 6.13 years) after obtaining a written informed consent. Data from two participants were excluded due to quality concerns (\autoref{subsec:eye_features}). All participants held at least a Master's degree from institutions where English was the primary language of instruction (high proficiency in English was a criterion to ensure accurate task comprehension and response quality). Participants had normal or corrected vision, and none reported neurological or psychiatric histories or medications.

\subsection{Task and procedure} 
\label{subsec:exp_setup}

\begin{figure}[h!]
    \centering
    \includegraphics[width=0.9\textwidth]{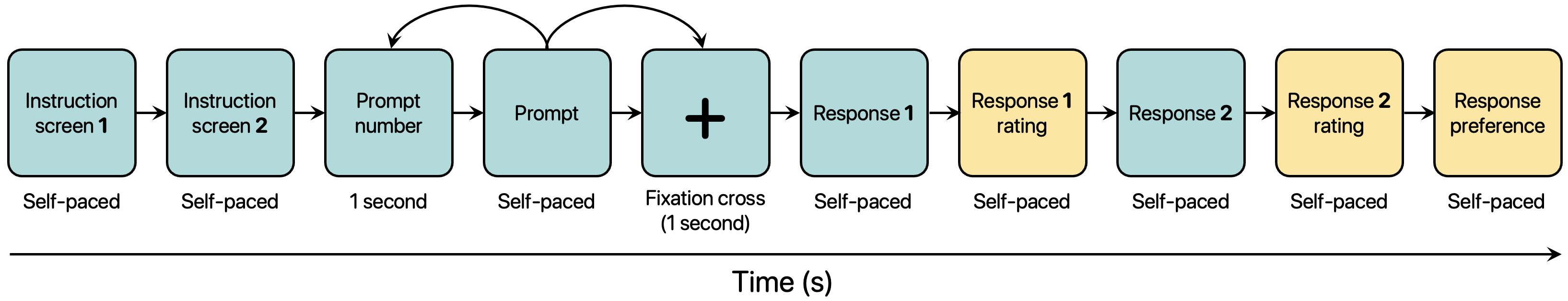}
    \caption{
    Task procedure: Tasks start with instructions. Participants see a prompt screen and choose to skip (up to 15 times) or answer. Choosing to answer leads to a fixation cross, followed by a rating for response 1, then response 2. Finally, participants select the preferred response, starting a new task.
    }
    \label{fig:task_procedure}
\end{figure}

Participants provided demographic information and acknowledged payment terms upon arrival at the lab. The experiment began with a detailed overview of the study, and researchers confirmed that each participant understood the instructions before proceeding. The experimental sequence, shown in \autoref{fig:task_procedure}, followed these steps for each prompt. We used PsychoPy \citep{peirce_psychopy2_2019} for stimulus presentation and behavioral response recording, given its flexibility and compatibility with various experimental hardware setups.

Specifically, the task required participants to annotate preferences by evaluating and selecting between two responses per prompt, following a streamlined version of the OASST1 annotation procedure. Participants reviewed guidelines (Appendix \ref{sec:app:corpus}) adapted from the original dataset's annotation protocol. Unlike other eye-tracking studies conducted in naturalistic reading settings, participants could choose not to annotate a response if they felt they lacked sufficient background knowledge. Each participant was asked to evaluate 30 prompts (from a set of 45) with two responses each, with the option to skip up to 15 prompts if they felt unqualified to assess them.

\subsection{Dataset description} \label{subsec:dataset_size}

We designed our experimental setup with a focus on balancing internal (prompt reading) and external validity (diversity of prompts). Recognizing the challenges that come with a limited dataset, we took a pragmatic approach by dividing our 360 prompts into 8 subset of 45.
Each such subset was assigned to three participants. Participants were required to respond to only 30 prompts (out of 45) within their assigned subgroup. Since participants could opt out of annotating prompts if they felt they lacked sufficient background knowledge, not all prompt-response pairs were annotated by all three participants. \autoref{tab:data_samples} shows the number of responses annotated by all participants (\textit{all}) and those with unanimous annotations across all three participants in each group and overall (\textit{unanimous annotated}). In cases where multiple participants read the same text, reading measures are averaged across participants to mitigate individual differences and yield a more consistent signal.

\begin{table}[h!]
    \centering
    \caption{Number of responses annotated in each subgroup.}
    \label{tab:data_samples}
    \begin{tabular}{c|ccccccccc}
    \toprule
    & \textbf{total} & \textbf{1} & \textbf{2} & \textbf{3} & \textbf{4} & \textbf{5} & \textbf{6} & \textbf{7} & \textbf{8} \\ 
    \hline
    \textbf{all} & 652 & 82 & 84 & 84 & 80 & 78 & 76 & 86 & 82 \\
    \textbf{unanimous annotated} & 214 & 30 & 46 & 20 & 22 & 24 & 36 & 22 & 14 \\
    \bottomrule
    \end{tabular}
\end{table}

\subsection{Apparatus} 
\label{subsec:data_adquisition}

For our study, we used a 27-inch monitor with dimensions 23.53 inches (width) by 13.24 inches (height). Eye movement data was collected using a GP3 HD Eye Tracker, sampling at 60 Hz, in accordance with recommended guidelines \cite{leiva2024modeling}. This equipment provides high spatial and temporal resolution, allowing for precise tracking of rapid eye movements \cite{cuve2022validation}. For each participant, we did a five-point calibration prior to the study, which was repeated until minimal error was achieved. Throughout the experiment, various eye movements -- including fixations, saccades, and blinks -- were recorded. The data included timestamps, gaze coordinates (x and y), and eye validity codes, with continuous monitoring of data quality through real-time feedback \cite{duchowski2017eye}.

\subsection{Data pre-processing} 
\label{subsec:eye_features}

The raw \acrshort{et} data (\autoref{subsec:data_adquisition}) was processed using the \textit{GazePoint Analysis} software \footnote{\url{https://www.gazept.com}}, which applies microsaccade detection algorithms to generate filtered fixation sequences. Each participant's data was stored in two formats: raw data in \textit{all\_gaze.csv} and filtered fixation sequences in \textit{fixations.csv}. For our fixation sequence analyses, we relied on \textit{fixations.csv}. However, we also include the raw data files to allow for other fixation detection approaches. Details on the eye tracker's features are available in the Open Gaze API manual from Gazepoint\footnote{Open Gaze API by Gazepoint: \url{https://www.gazept.com/dl/Gazepoint_API_v2.0.pdf}}. Each fixation data entry includes its sequential order, pixel coordinates (x, y) on the display, pupil dilation per eye, and fixation duration in seconds.

\begin{figure}[h!]
    \centering
     \includegraphics[width=0.65\linewidth]{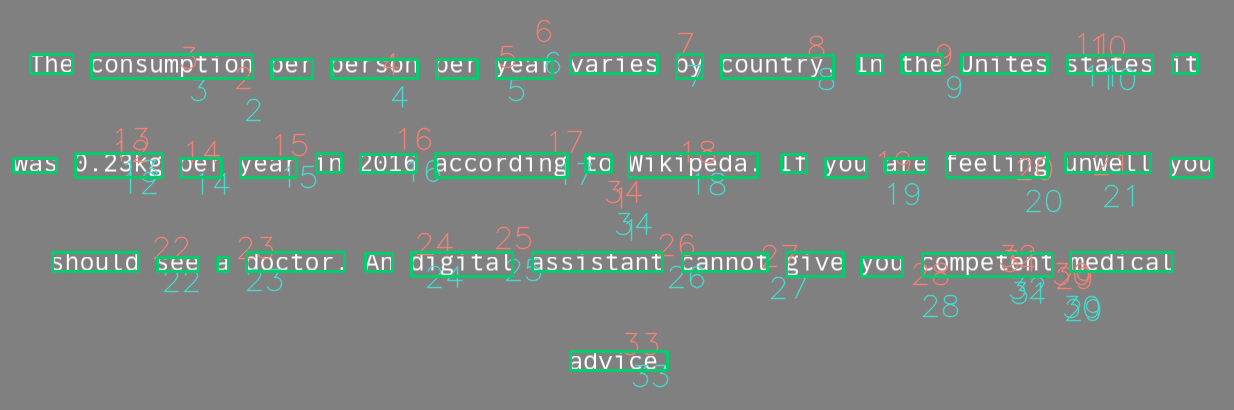}
    \caption{Example of a cropped screenshot of the original image with \acrshort{aoi} shown in green boxes. The original fixation sequence is indicated by blue numbers (suggesting viewing order), while calibrated fixations are displayed by pink numbers.}
    \label{fig:example_fixations}
\end{figure}

Fixations for each participant's response were aligned with individual words by capturing screenshots via \textit{PsychoPy} \cite{peirce_psychopy2_2019}. Using \textit{OpenCV}'s \acrfull{ocr} capabilities, we extracted the coordinates for each \acrfull{aoi} (bottom-left position, width, and height). Each area represents a single word, defined by spaces, with punctuation typically included with adjacent words (\autoref{fig:example_fixations}). Fixations were matched to the nearest \acrshort{aoi} through Euclidean distance calculations, with several key adjustments. Following prior work \cite{hollenstein_zuco_2018, bolliger_emtec_2024}, a maximum distance threshold was applied to exclude fixations that fell outside the expected reading area, such as screen edges or corners. Initial fixations were also excluded, as manual checks revealed that they frequently centred on the screen's cross marker, where participants typically start before shifting to the text.

A common challenge in eye-tracking is the vertical drift of recorded gaze coordinates, where fixations may be incorrectly attributed to text lines above or below the intended reading line \cite{carr_algorithms_2022, bolliger_emtec_2024}. To mitigate this drift in left-to-right reading and ensure accurate fixation-to-word mapping, we applied the following correction: if a fixation's nearest \acrshort{aoi} appeared on a different line than the previous fixation, but to the right along the x-axis, we identified this as vertical drift. In those cases, the fixation was reassigned to the closest word on the same line as the previous fixation's area.

Before reading each response, participants focused on an on-screen cross (\autoref{fig:task_procedure}) that was used as a calibration reference point. By comparing the actual position of the cross to participant's gaze location, we adjusted subsequent fixations to account for any misalignment. For each response, we calculated the mean fixation-to-word distance with and without calibration and used the approach that minimized this average distance (\autoref{fig:example_fixations}). When neither method yielded distances below our threshold, the response was excluded. This approach led to the exclusion of two participants (with five and two responses excluded, respectively), requiring additional sessions to collect complete data with replacement participants. 

\begin{table}[t!]
    \caption{Means and standard deviations for reading measures across subgroups (n=3) and all participants (n=24)}
    \centering
    \begin{adjustbox}{width=0.9\textwidth}
    \begin{tabular}{llllllll}
    \toprule
    Subgroup & Total Words & Word Length & TRT (s) & Omission Rate (\%) & nFix & FFD (s) & Pupil Size (pixel) \\
    \midrule
    1 & 59.68 ($\pm$ 34.38) & 4.77 ($\pm$ 0.74) & 0.49 ($\pm$ 0.12) & 0.48 ($\pm$ 0.14) & 0.74 ($\pm$ 0.30) & 0.35 ($\pm$ 0.05) & 12.44 ($\pm$ 1.49) \\
    2 & 50.88 ($\pm$ 35.57) & 4.74 ($\pm$ 0.83) & 0.60 ($\pm$ 0.53) & 0.52 ($\pm$ 0.16) & 0.89 ($\pm$ 0.90) & 0.34 ($\pm$ 0.07) & 13.32 ($\pm$ 2.19) \\
    3 & 50.90 ($\pm$ 33.57) & 4.87 ($\pm$ 0.76) & 0.50 ($\pm$ 0.37) & 0.52 ($\pm$ 0.14) & 0.77 ($\pm$ 0.69) & 0.32 ($\pm$ 0.04) & 13.83 ($\pm$ 1.97) \\
    4 & 56.79 ($\pm$ 34.35) & 4.68 ($\pm$ 0.75) & 0.57 ($\pm$ 0.28) & 0.49 ($\pm$ 0.15) & 0.82 ($\pm$ 0.57) & 0.38 ($\pm$ 0.08) & 18.16 ($\pm$ 1.40) \\
    5 & 61.87 ($\pm$ 34.97) & 4.80 ($\pm$ 0.72) & 0.49 ($\pm$ 0.21) & 0.53 ($\pm$ 0.14) & 0.73 ($\pm$ 0.48) & 0.34 ($\pm$ 0.06) & 18.37 ($\pm$ 3.51) \\
    6 & 68.20 ($\pm$ 36.11) & 4.92 ($\pm$ 0.70) & 0.47 ($\pm$ 0.22) & 0.54 ($\pm$ 0.15) & 0.77 ($\pm$ 0.53) & 0.30 ($\pm$ 0.04) & 12.93 ($\pm$ 3.43) \\
    7 & 59.29 ($\pm$ 33.48) & 4.84 ($\pm$ 0.76) & 0.46 ($\pm$ 0.19) & 0.54 ($\pm$ 0.09) & 0.69 ($\pm$ 0.38) & 0.31 ($\pm$ 0.03) & 12.14 ($\pm$ 3.93) \\
    8 & 55.96 ($\pm$ 33.44) & 4.86 ($\pm$ 0.91) & 0.65 ($\pm$ 0.70) & 0.52 ($\pm$ 0.12) & 0.75 ($\pm$ 0.64) & 0.39 ($\pm$ 0.08) & 14.71 ($\pm$ 2.56) \\
    \midrule
    Total & 57.82 ($\pm$ 34.94) & 4.81 ($\pm$ 0.78) & 0.53 ($\pm$ 0.39) & 0.51 ($\pm$ 0.14) & 0.78 ($\pm$ 0.60) & 0.34 ($\pm$ 0.07) & 14.51 ($\pm$ 3.46) \\
    \bottomrule 
    \end{tabular}
    \end{adjustbox}
    \label{tab:reading_measures}
\end{table}

\subsection{Descriptive statistics on reading measures}
\label{sec:compute_readingm}

For each response, we computed the total word count and average word length, along with several additional reading measures that capture participants' reading behaviours (\autoref{tab:reading_measures}). Drawing from previous eye-tracking corpora we calculated measures such as the total number of fixations (nFix), to assess basic word-level engagement, and first fixation duration (FFD), to reflect early lexical and syntactic processing \citep{hollenstein_multilingual_2021}. For indicators of later syntactic processing and disambiguation \citep{hollenstein_multilingual_2021} we included \acrshort{trt} on words receiving multiple fixations. 


Word count and average word length were analysed for both response types (chosen vs. rejected). Responses in the chosen category had significantly higher word counts (\textit{t}(23) = 16.53, \textit{p} < .001, \textit{d} = 3.37) and greater average word lengths (\textit{t}(23) = 3.10, \textit{p} < .001, \textit{d} = 0.63) compared to the rejected responses (\autoref{fig:words_lenght}). Moreover, we observed that reading measures correlated strongly with word length (\autoref{tab:corre_reading_measures}), indicating that longer words naturally require additional processing time as readers scan their entire character span. These results align with prior findings linking reading behaviour to both word length and word frequency \citep{hollenstein_relative_2021, haller_measurement_2024}.
   

~\linebreak

\begin{minipage}{0.55\textwidth}
    \centering
    \includegraphics[width=0.8\linewidth]{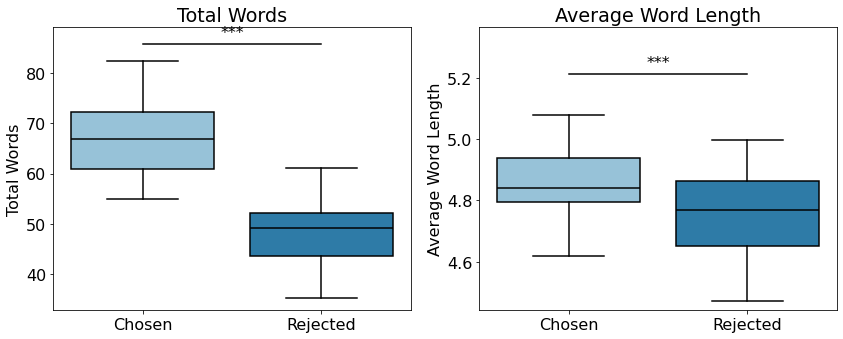}
    \captionof{figure}{Words amount and length by condition}
    \label{fig:words_lenght}
\end{minipage}
\hfill
\begin{minipage}{0.43\textwidth}
    \centering
    \captionof{table}{Correlation between word length and reading measures.}
    \begin{adjustbox}{width=0.95\textwidth}
        \begin{tabular}{llll}
            \toprule
            & TRT & FFD & nFix \\
            \midrule
            All responses & 0.43 ($\pm$ 0.21) & 0.39 ($\pm$ 0.22) & 0.46 ($\pm$ 0.22) \\
            Unanimous response & 0.49 ($\pm$ 0.19) & 0.44 ($\pm$ 0.21)& 0.52 ($\pm$ 0.19) \\
            \bottomrule
        \end{tabular}
    \end{adjustbox}
    \label{tab:corre_reading_measures}
\end{minipage}

%% file: 05_analyse.tex
\section{Preference analysis with organic vs. synthetic reading measures} 
\label{sec:analyse_reading}

\subsection{Methodology}
We compare the reading measures of number of fixations (nFix), total reading time (TRT), and first fixation duration (FFD) between preferred and rejected responses. For labeling preferred and rejected responses, we retain the original dataset's labels (Appendix \ref{app:annotations}), as these would be used for the \acrshort{llm} alignment with this data and also to ensure reproducibility across studies. To investigate whether similar patterns are present in synthetic reading measures, we replicate the study using reading measures produced by a generative model that processes each response independently as input. Our analysis considers both all responses, and only unanimous responses (\autoref{tab:data_samples}).

\noindent\textbf{Synthetic Reading Measures:} We apply a RoBERTa-based model \citep{li_torontocl_2021}, previously used in studies such as \cite{zhang_eye-tracking_2024} and \cite{wang_gaze-infused_2024}, to generate synthetic reading measures. The model uses a regression head on each token with a linear layer to predict five measures: FFD, fixProp, GPT, TRT, and nFix (\autoref{tab:et-features}). This model is initialized with pre-trained weights and fine-tuned on the ZUCO1 \citep{hollenstein_zuco_2018}, ZUCO2 \citep{hollenstein_towards_2020}, and PROVO \citep{luke_provo_2018} datasets, using 800 sentences (15.7 tokens per sentence), with 191 sentences (3.5k tokens) held out for evaluation. During training, word-level features are mapped to token-level features by assigning them to the first token of each word, with the remaining tokens given zero values. For inference, this process is reversed to obtain word-level features. Due to potential mismatches between tokenized words and OCR-extracted words, we map one word list to another; if one OCR word corresponds to multiple tokenized words, their reading measures are summed (details in Appendix \ref{sec:app:mapping_features_syn}).

\subsection{Results}

\begin{figure}[h!]
    \centering
     \includegraphics[width=0.9\linewidth]{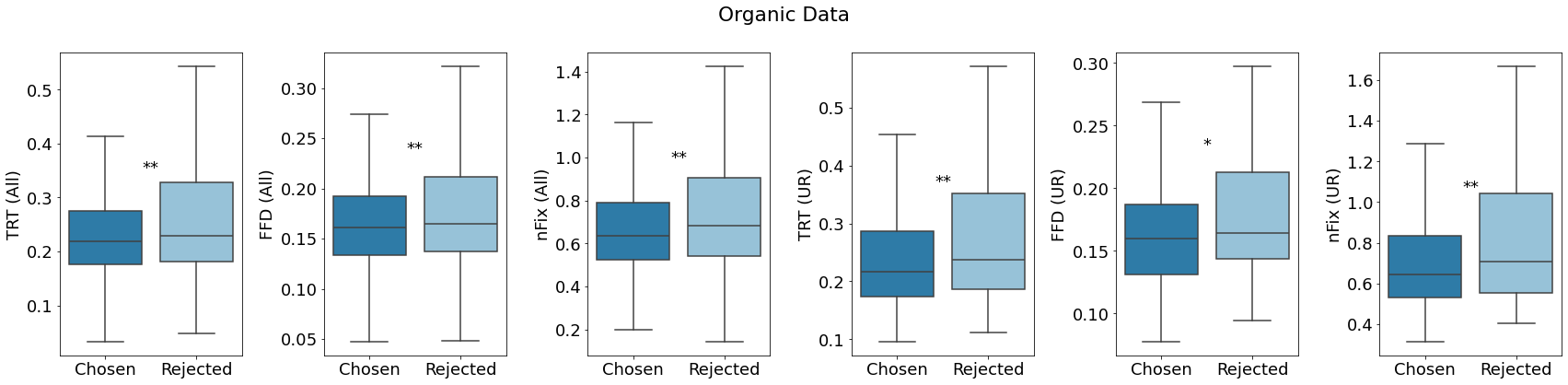}
    \caption{Reading measures comparison between responses on organic \acrshort{et} corpus for All (AR) and Unanimous Responses (UR).}
    \label{fig:reading_measure_real}
\end{figure}

\begin{figure}[h!]
    \centering
     \includegraphics[width=0.9\linewidth]{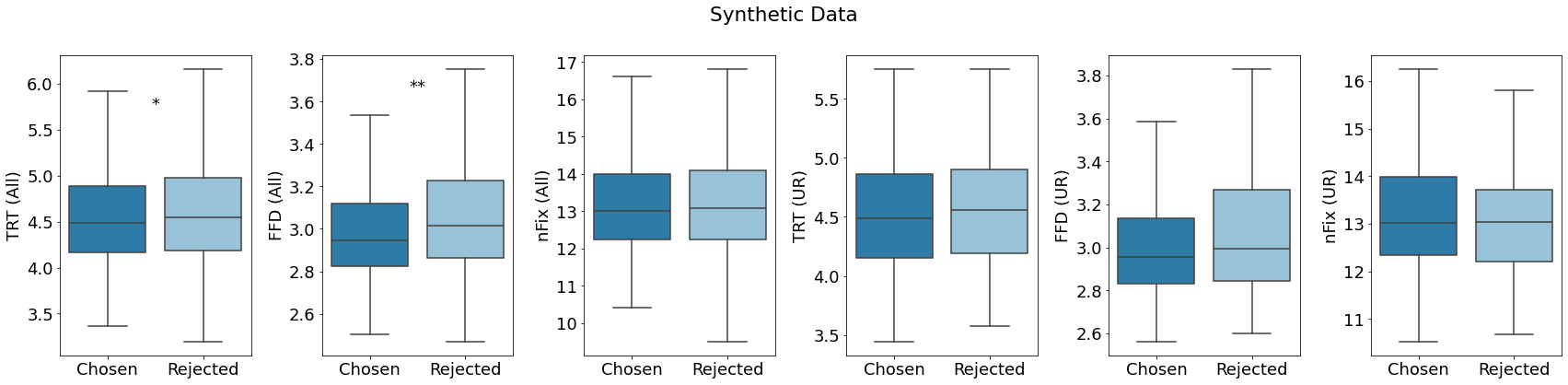}
    \caption{Reading measures comparison between responses on synthetic \acrshort{et} corpus for All (AR) and Unanimous Responses (UR).}
    \label{fig:reading_measure_synthe}
\end{figure}

When analysing all responses (\autoref{fig:reading_measure_real}), significant differences were observed across all metrics, possibly influenced by correlations shown in \autoref{tab:corre_reading_measures}. Rejected responses showed significantly higher TRT ($t(325)=-3.76$, $p < .01$, $d=-.20$), higher FFD ($t(325)=-3.37$, $p < .01$, $d=-.18$), and higher nFix ($t(325)=-3.21$, $p < .01$, $d=-.17$) than preferred responses. An analysis of unanimous responses revealed similar effects, with rejected responses having higher TRT ($t(106)=-2.72$, $p < .01$, $d=-.26$), higher FFD ($t(106)=-2.07$, $p < .05$, $d=-.20$), and higher nFix ($t(106)=-2.63$, $p < .01$, $d=-.25$) (\autoref{fig:reading_measure_real}).

When analysing synthetic data generated by fixation model (\autoref{fig:reading_measure_synthe}), TRT ($t(321)=-2.27$, $p < .05$, $d=-.12$) and FFD ($t(321)=-4.05$, $p < .01$, $d=-.22$) differed significantly across all responses, though nFix showed no significant difference ($t(321)=-1.20$, $p=.22$, $d=-.06$). Moreover, unanimous responses showed no significant differences for TRT ($t(106)=-1.00$, $p=.31$, $d=-.09$), FFD ($t(106)=-1.97$, $p=.051$, $d=-.19$), or nFix ($t(106)=-0.16$, $p=.87$, $d=-.01$).

\begin{figure}[h!]
    \centering
    \begin{subfigure}[c]{0.24\textwidth}
        \includegraphics[width=1\textwidth]{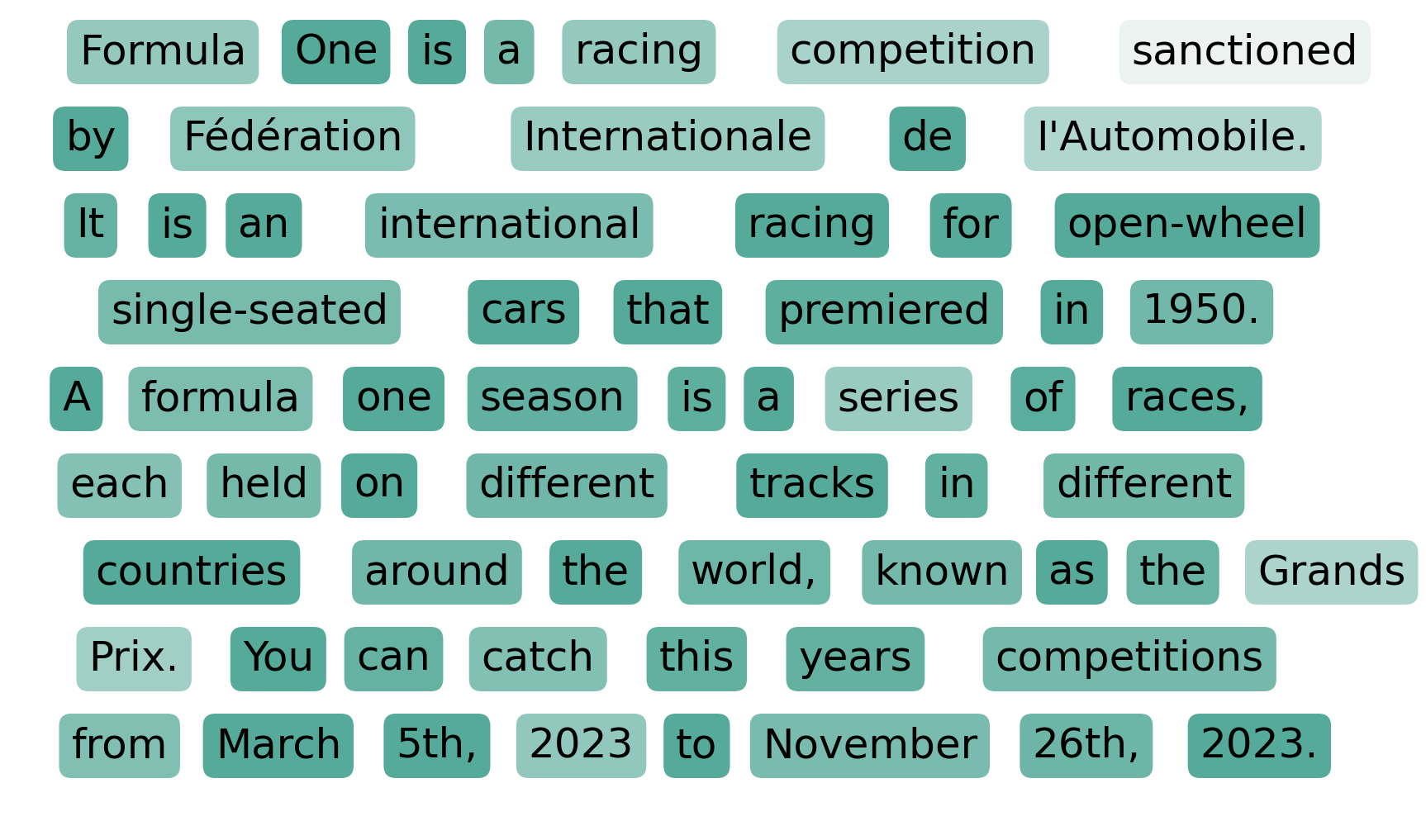}
        \label{fig_alg_a1}
    \end{subfigure}
  \begin{subfigure}[c]{0.24\textwidth}
        \includegraphics[width=1\textwidth]{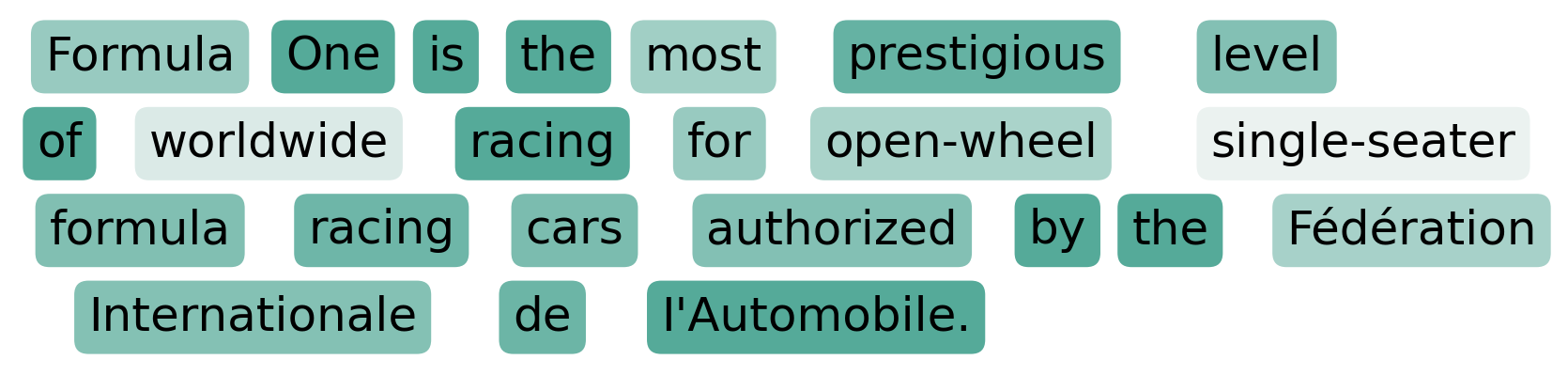}
        \label{fig_alg_c1}
    \end{subfigure}
    \begin{subfigure}[c]{0.24\textwidth}
        \includegraphics[width=1\textwidth]{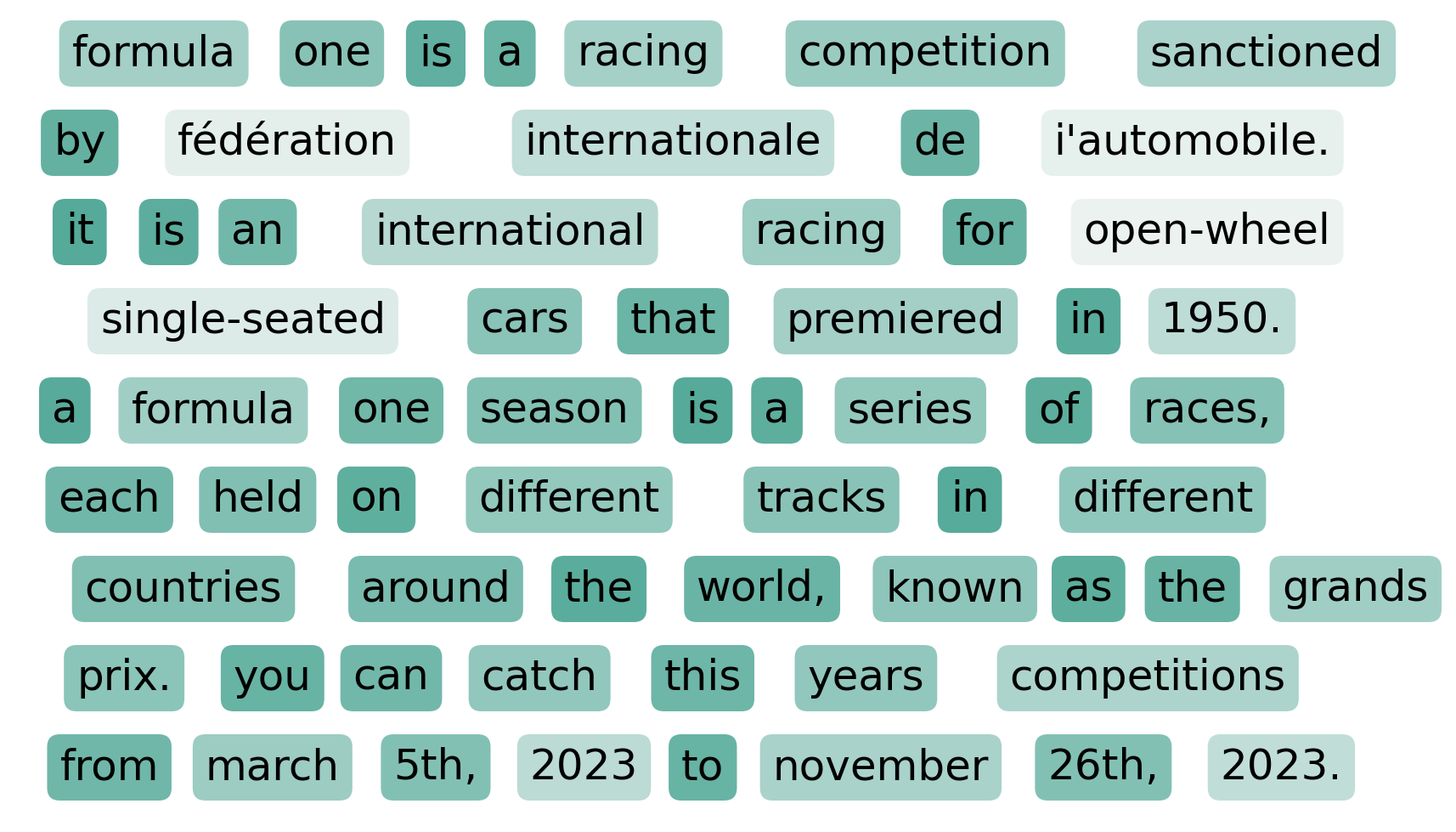}
        \label{fig_alg_b1}
    \end{subfigure}
    \begin{subfigure}[c]{0.24\textwidth}
        \includegraphics[width=1\textwidth]{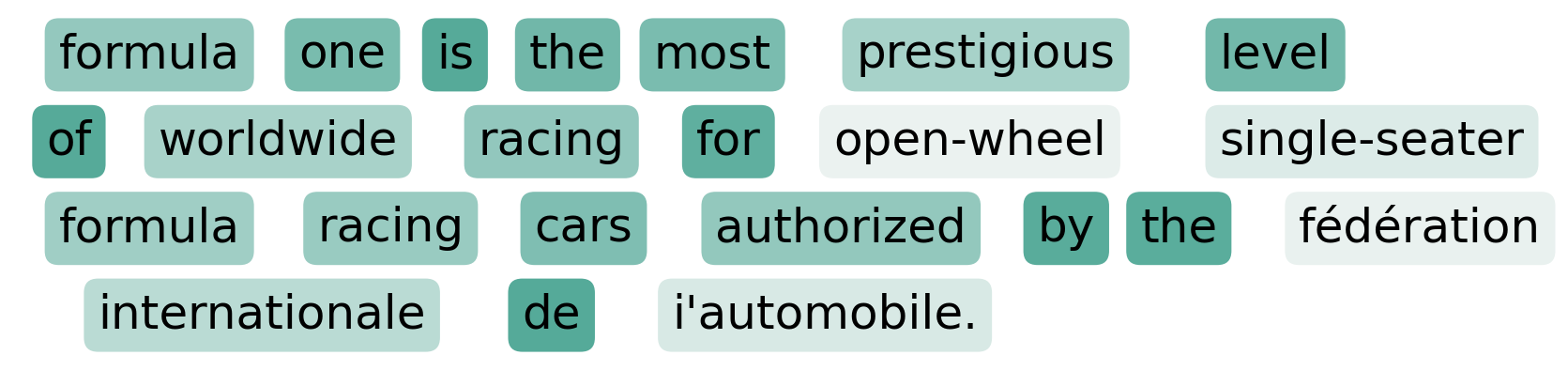}
        \label{fig_alg_d1}
    \end{subfigure}
    \caption{ Word-by-word reading time comparison between real (left) and synthetic (right) eye tracking data for the question "What is Formula One?". The first and third plots display preferred responses, while the second and fourth show rejected ones, with darker shades indicating longer fixation durations.}
    \label{fig:trt_example}
\end{figure}

\autoref{fig:trt_example} shows a comparison between representative TRT patterns for preferred versus rejected responses. Differences in TRT distributions between organic and synthetic data suggest that high TRT values vary noticeably between the two. This difference could be attributed to several factors, such as the relatively small dataset that the model was trained on, thus limiting its vocabulary and leading to a possible mismatch with the target words \citep{huang_longer_2023}. Furthermore, the generative model used here lacks task-specific data, preventing it from optimizing predicted measures for this specific task -- a limitation noted in other studies \cite{huang_long-range_2023, huang_longer_2023}. When comparing reading measures for chosen and rejected responses, between the organic (\autoref{fig:reading_measure_real}) and synthetic (\autoref{fig:reading_measure_synthe}) \acrshort{et} data, the differences observed in the latter are less pronounced, as expected given the lack of task-specific training. Research suggests that fixation prediction models could benefit from task-specific corpora or multi-task learning to enhance their performance \cite{sood_improving_2020,prasse_sp-eyegan_2023}, which indicates the potential value of this dataset in improving generative models for human-aligned NLP tasks.

%% file: 06_analyse_attention.tex
\section{Human- vs. model-based attention measures}
\subsection{Methodology}

Considering our \acrshort{et} corpora and language models, we analyse token-level relative importance for each prompt response \( \mathbf{x} = \langle x_1, \dots, x_N \rangle \) , where \( N \) is the word count and \( x_j \) denotes the \( j \)-th word. For each response, we derive two importance vectors: a human-assigned importance vector \( \mathbf{h} = \langle h_1, \dots, h_N \rangle \) and a model-assigned importance vector \( \mathbf{m} = \langle m_1, \dots, m_N \rangle \), where \( h_j \) and \( m_j \) indicate the importance assigned to word \( x_j \) by human readers and by the model, respectively. We quantify the alignment between these human and model importance values by calculating the mean Spearman correlation across all responses \( \mathbf{x} \), following \citet{eberle_transformer_2022, hollenstein_relative_2021}.

\noindent\textbf{Human-based attention:} Token-level importance to human readers is closely correlated with fixation duration \cite{hollenstein_relative_2021}. Generally, the more frequently a token is fixated upon, the higher its importance in answering the prompt question \cite{sood_interpreting_2020}. Previous studies \cite{hollenstein_relative_2021, wu_eye_2024, bensemann_eye_2022} have used TRT to measure relative attention per token, though alternative reading metrics have also been examined \cite{wang_gaze-infused_2024, wang_probing_2024}. In our study, in addition to TRT, we compute FFD and nFix per word. TRT and FFD are normalized by response, where fixation durations for each token \( x_j \) are expressed as \( d_{j,r} = \frac{t_{j,r}}{\sum_j t_{j,r}} \), as done in \cite{hollenstein_relative_2021, wu_eye_2024}.

\noindent\textbf{Model-based attention:} Following \cite{bensemann_eye_2022, wang_gaze-infused_2024, hollenstein_relative_2021}, we investigate Transformer-based language models with varying architectures, differentiated by their base module, number of layers, and attention heads. Each response is independently tokenized and processed through the model to extract attention matrices from each attention head. Token-level attention values are derived from pre-softmax output weights before interaction with $V$ (as shown in \autoref{eq:attention}), averaging token attention across heads within each layer and aggregating attention received by each token. The resulting vector is normalized within each sentence to indicate the proportional attention each token receives from others in the response. This process remains consistent across models, regardless of their number of layers and heads. Each model uses its specific tokenizer and special tokens (\textit{[CLS], [SEP], [INS]...}), though attention values for special tokens are omitted from these calculations. Final attention values are normalized per sentence to show the proportional attention given to each word. Token-level attention values are then converted to word-level attention by aggregating them.  Human attention is focused on words obtained through OCR (\autoref{subsec:data_adquisition}), which may not match those from the tokenizer. Therefore, we map and add the attention from the tokenizer's word list that corresponds to an OCR word (detailed in Appendix \ref{sec:app:mapping_features_att}).

\begin{equation}
\text{Attention}(Q, K, V) = \text{softmax}\left(\frac{QK^\top}{\sqrt{n}}\right) V
\label{eq:attention}
\end{equation}

\subsection{Experimental setup}
\noindent\textbf{Datasets.} We use our \acrshort{et} dataset for this analysis. For each trial, we calculate a correlation score between human- and model-based attention, reporting both the average correlation across all trials and unanimous responses (\autoref{tab:data_samples}).


\noindent\textbf{Models.} We include both encoder-based and decoder-based Transformer models, detailed in \autoref{table:models} and Appendix \ref{sec:app:atenttion_models}. Pretrained models from Huggingface\footnote{\url{https://huggingface.co}} are used directly via their checkpoint names, with minimal modifications for \acrshort{rms} to obtain attention patterns (see Appendix \ref{sec:app:atenttion_models}). Information regarding the hardware used in Appendix \ref{sec:hadware}.


\begin{table}[h!]
    \centering
    \caption{Models used in each category}
    \begin{adjustbox}{width=0.75\textwidth}
    \begin{tabular}{lp{10cm}}
    \toprule
    \textbf{Category} & \textbf{Models} \\
    \midrule
    Masked \acrshort{llms} & BERT \cite{devlin_bert_2019} (bert-base-uncased, bert-base-cased, bert-large-uncased), RoBERTa \cite{liu_roberta_2019} (roberta-base, roberta-large) \\
    \cmidrule(lr){1-2}
    Pretrained \acrshort{llms} & Phi 1.5 \cite{li_textbooks_2023} (phi-1\_5), Llama 2 \cite{touvron_llama_2023} (Llama-2-7b-hf), Llama 3 \cite{dubey_llama_2024} (Meta-Llama-3-8B, Llama-3.1-8B), Mistral \cite{jiang_mistral_2023} (Mistral-7B-v0.1) \\
    \cmidrule(lr){1-2}
    Human aligned \acrshort{llms} & Llama 2 \cite{touvron_llama_2023} (Llama-2-7b-chat-hf), Llama 3 \cite{dubey_llama_2024} (Meta-Llama-3-8B-Instruct, Llama-3.1-8B-Instruct), Mistral \cite{jiang_mistral_2023} (Mistral-7B-Instruct-v0.1) \\
    \cmidrule(lr){1-2}
    Reward Models & UltraRM \cite{cui_ultrafeedback_2024} (UltraRM-13b), Eurus \cite{yuan_advancing_2024} (Eurus-RM-7b), QRM \cite{dorka_quantile_2024} (QRM-Llama3.1-8B) \\
    \bottomrule
    
    \label{table:models}
    \end{tabular}
    \end{adjustbox}
\end{table}

\noindent\textbf{Transformer encoder-based models.} We use several versions of BERT \citep{devlin_bert_2019} trained on Masked Language Modeling (MLM) and Next Sentence Prediction (NSP) tasks, along with two variations of RoBERTa \citep{liu_roberta_2019}. While RoBERTa shares BERT's core architecture, it enhances the training approach. Encoder-based models excel in contextualizing each token in both directions.

\noindent\textbf{Transformer decoder-based models.} Decoder models, while less frequently analysed in attention alignment studies, are widely used in generation-oriented language models, offering insights into how alignment holds in text-generation tasks. We consider several  state-of-the-art, open-source models, such as Llama \citep{touvron_llama_2023, dubey_llama_2024} and Mistral \citep{jiang_mistral_2023}, and the compact model Phi 1.5 \citep{li_textbooks_2023}, to explore the impact of model scale on alignment. Additionally, three \acrlong{rms} are evaluated for their alignment capabilities on this specific task, as discussed in \autoref{sec:backgroun_llms}.

\subsection{Results}

\subsubsection{Correlation analysis of model layers and reading measures} 
\label{sec:analuse_layers}

\begin{figure}[h!]
    \centering
    \subfloat{
        \includegraphics[width=0.23\textwidth]{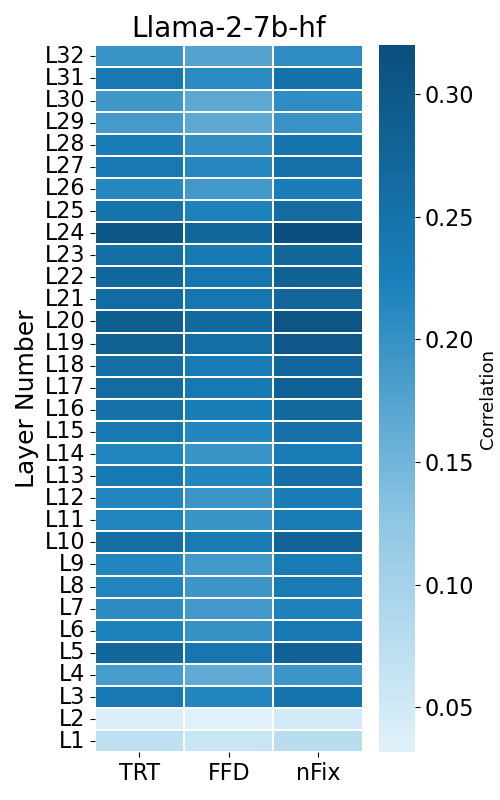}
        \label{fig_att_layer_1}
    }
    \subfloat{
        \includegraphics[width=0.23\textwidth]{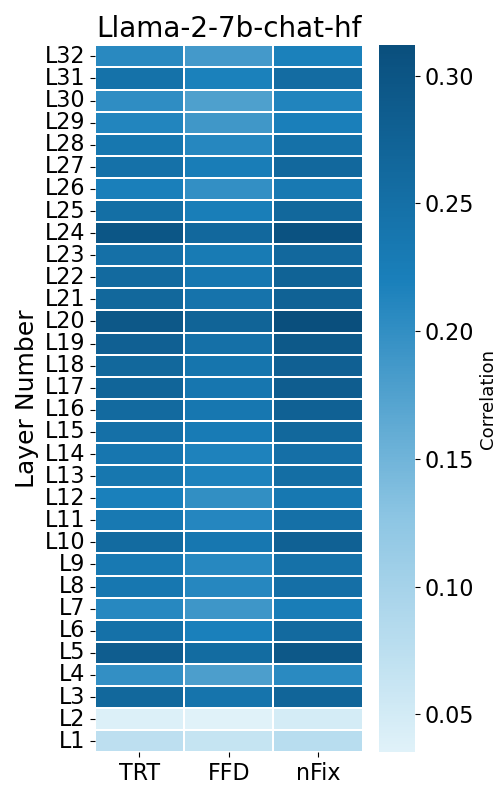}
        \label{fig_att_layer_2}
    }
    \subfloat{
        \includegraphics[width=0.23\textwidth]{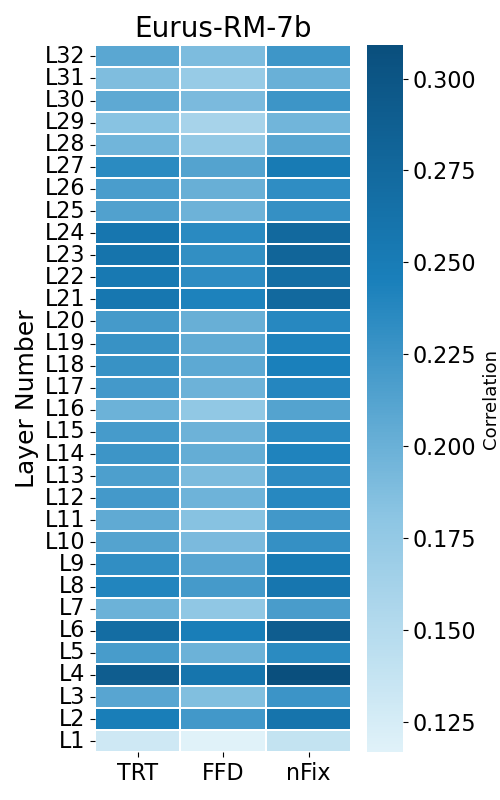}
        \label{fig_att_layer_3}
    }
    \subfloat{
        \includegraphics[width=0.23\textwidth]{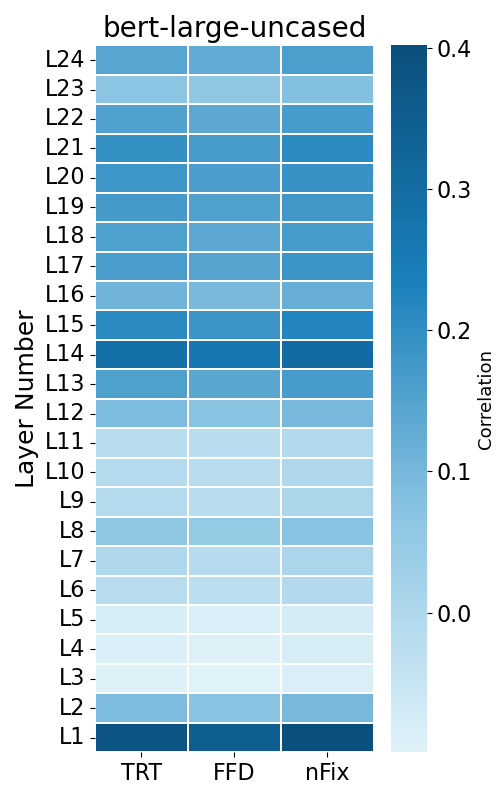}
        \label{fig_att_layer_4}
    }
   
    \caption{Mean Spearman correlation values of different layers in different models with TRT, FFD and nFix reading measures.}
     \label{fig:att_layer}
\end{figure}

\autoref{fig:att_layer} shows the correlation of each layer with the reading measures TRT, FFD, and nFix across representative models from each family (detailed results for all models are available in Appendix \ref{app:sec:analuse_layers}). Across all reading measures, FFD consistently demonstrates the weakest correlation across all architectures. Encoder-based models show the highest correlation in the first layer and also a significant correlation in the middle layers when compared to decoder-based models. In contrast, decoder-based models typically exhibit low correlation in the initial layers, with the exception of Phi, likely due to its smaller size. The strongest correlations are observed in the middle and upper layers, consistent with findings reported by \citet{wang_probing_2024}, indicating that these layers are more attuned to human attention patterns, particularly in recognizing contextually significant tokens. Among the various model variants, both original and aligned models generally display similar layer-wise distributions, as seen with Llama 2 and its Chat variant. While some research (e.g., \cite{bensemann_eye_2022}) focuses on the first-layer encoding, others (e.g., \cite{hollenstein_relative_2021, sood_interpreting_2020}) highlight the final layer. For our subsequent analyses, we selected each model's highest-correlating layer, recognizing that optimal alignment to human attention varies by architecture and task.

\subsubsection{Correlation analysis of model architectures and reading measures} 
\label{sec:analyse_models}

We analyse correlations between model attention patterns and human reading behaviours across trials, distinguishing between preferred and rejected responses. Specifically, correlations were observed across all three reading measures (FFD, TRT, and nFix), with TRT and nFix showing the strongest alignment across models (\autoref{fig_att_model}). In the Appendix \ref{app:sec:analyse_models}, \autoref{fig_att_model_TRT_appendix}, \autoref{fig_att_model_FFD_appendix}, and \autoref{fig_att_model_nfix_appendix} indicate that correlations were stronger for unanimous trials (left) than for all (right), likely due to reduced variability from data averaging and the smaller sample size.

\noindent\textbf{Model architecture.} Models with bidirectional encoder architectures consistently outperformed decoder-based models, despite not being specifically trained on tasks related to our dataset. This advantage may stem from the bidirectional processing of context, which more closely resembles human reading. Among encoder-based models, BERT-based architectures exhibited stronger correlations than RoBERTa variants, supporting findings from prior studies \cite{bensemann_eye_2022, wu_eye_2024}. Model size, however, had minimal impact on correlation strength.

\begin{figure}[h!]
    \centering
     \subfloat{\includegraphics[width=0.5\textwidth]{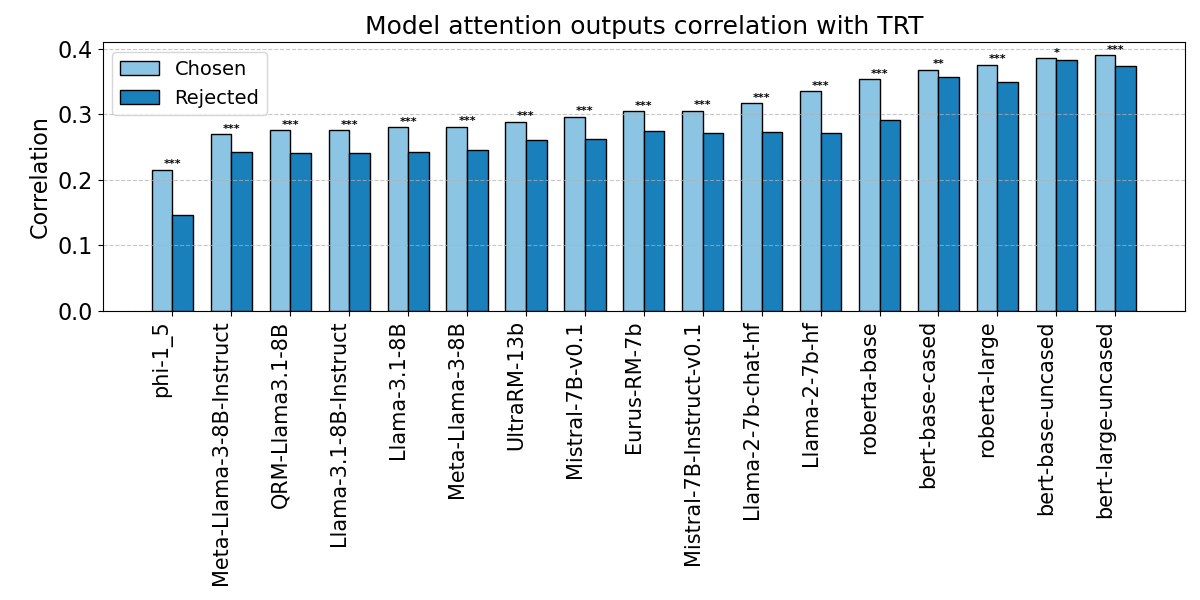}
    \label{fig_att_model_TRT_c}}
     \subfloat{\includegraphics[width=0.5\textwidth]{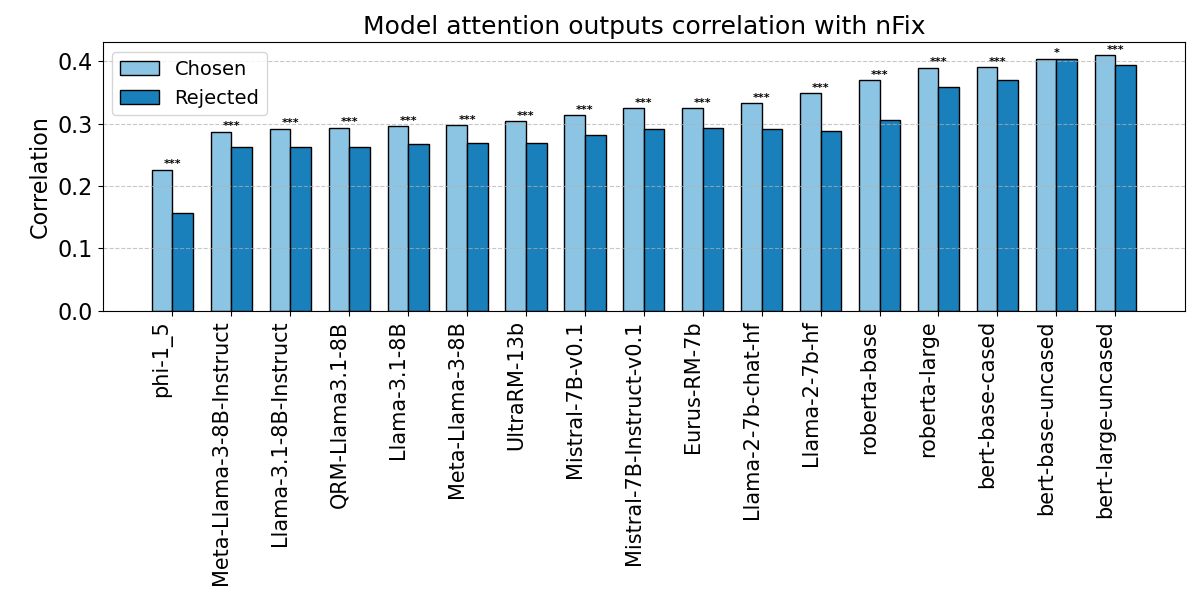}}
    \label{fig_att_model_TRT_all}
    \caption{Mean Spearman correlation analysis between TRT (left barplot) and nFix (right barplot) on the unanimous annotated dataset. (*) indicates statistical significance between chosen and rejected.}
    \label{fig_att_model}
\end{figure} 

\noindent\textbf{Model task.} Contrary to expectations, task-specific alignment did not consistently improve correlation strength. The standard, pretrained Llama 3.1 model, which lacked alignment or instruction-following tuning, generally achieved higher correlations than its instruction-tuned (Llama 3.1 Instruct) and \acrshort{rm} (QRM-Llama 3.1) counterparts. This suggests that task-specific alignment and reward tuning may not be essential for achieving strong human-like attention alignment, though other studies have highlighted the importance of such alignment \citep{wu_eye_2024}. Interestingly, decoder-based models demonstrated greater differentiation between chosen and rejected responses, with some BERT-based models showing higher average correlations for rejected responses. This trend indicates that task alignment has a greater influence on the differences in correlation between chosen and rejected responses than on the overall correlation itself. Additionally, newer Llama versions (Llama 3 and Llama 3.1), despite being trained on larger datasets, exhibited lower correlations across all reading measures. Other studies have also noted that models with higher performance can sometimes exhibit lower attention correlations \citep{wu_eye_2024}.

\begin{figure}[h!]
    \centering
    \subfloat{\includegraphics[width=0.33\textwidth]{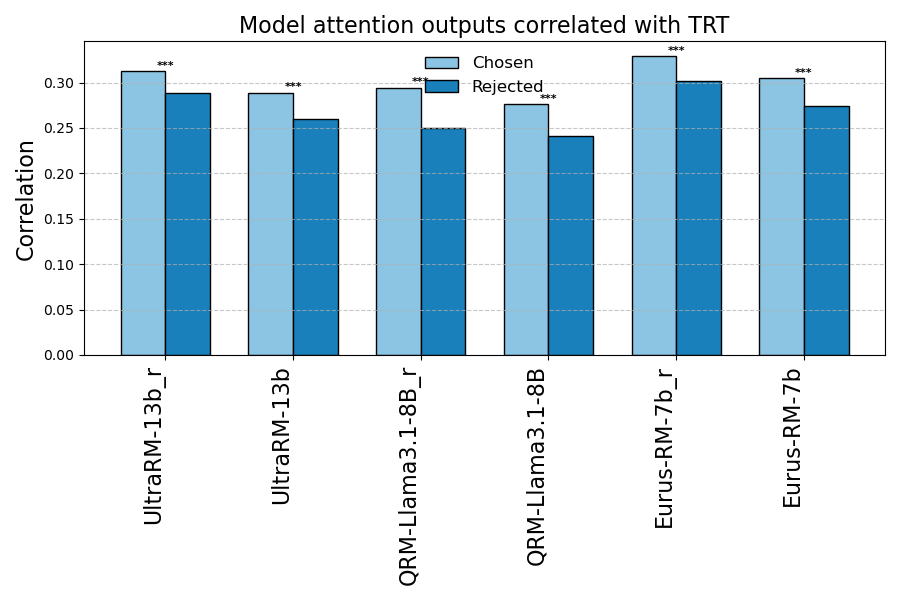}}
    \label{fig_att_model_trt_reward}
     \subfloat{\includegraphics[width=0.33\textwidth]{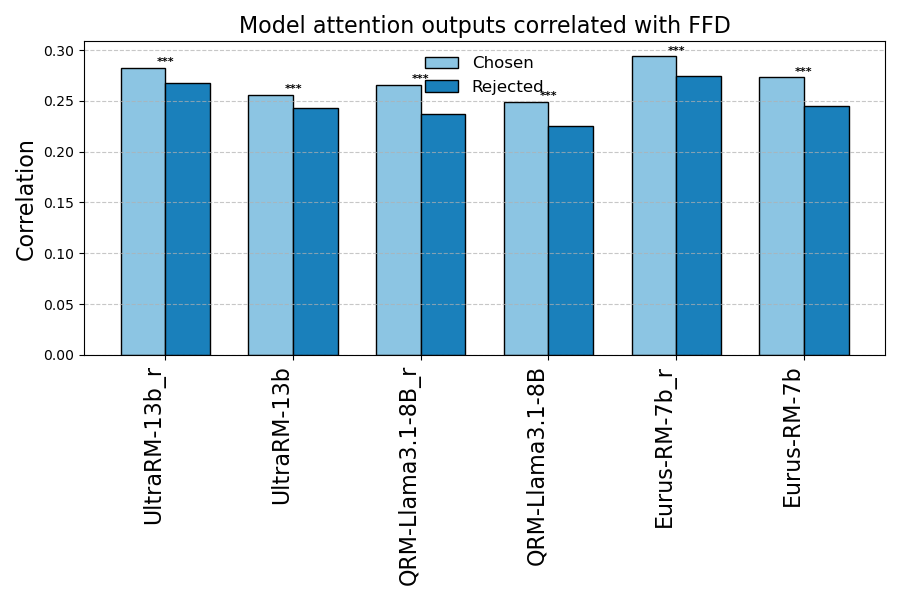}}
    \label{fig_att_model_ffd_reward}
    \subfloat{\includegraphics[width=0.33\textwidth]{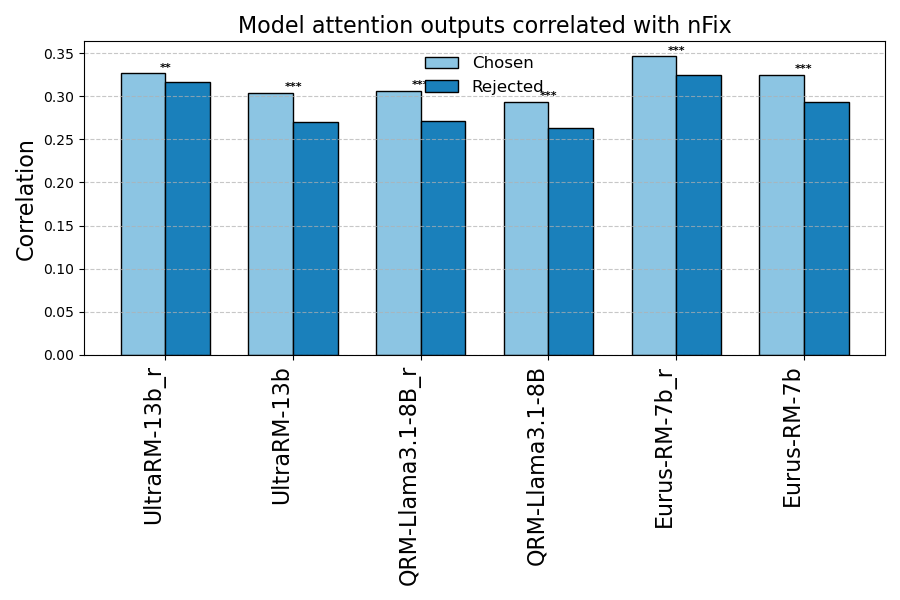}}
    \caption{Mean Spearman correlation coefficients comparing TRT, FFD, nFix across three Reward Models, with both standard and prompt-conditioned (\_r) variations.}
    \label{fig_att_model_reward}
\end{figure}

\subsubsection{Correlation analysis of \acrshort{rm} input and reading measures}
\label{sec:reward_modelling}

To examine task alignment more closely, we compared correlation results from the three \acrshort{rms} analysed previously. Our comparison used two input configurations: (1) isolated responses (as in previous analyses), and (2) concatenated prompt-response pairs (implementation details are in Appendix \ref{sec_app_reward_model}). Results, presented in \autoref{fig_att_model_reward}, suggest that models achieved stronger alignment with human attention patterns when processing the full prompt-response context. This setup aligns with human information processing, where the prompt is read before the response (see additional results in Appendix \ref{app:sec:reward_modelling}).

%% file: 07_conclusions.tex
\section{Discussion and conclusions}

This work introduced the first eye-tracking corpus specifically designed to study human alignment in \acrshort{llms}. Building upon a widely used dataset in state-of-the-art \acrshort{llms}, we performed a comprehensive analysis of the raw \acrshort{et} data and fixation sequences, and calculated reading measures to examine how humans process preferred versus rejected responses to identical prompts. Our findings revealed significant distinctions between these two response categories. When comparing these differences with the reading measures obtained from a generative model not specifically trained on \acrshort{et} corpora for this task, we observed similar but less pronounced differences, confirming the importance of task-specific alignment in model development.

Our investigation also explored the relationship between human reading measures and attention patterns across various Transformer-based architectures. This comparison highlighted differences in chosen and rejected responses and, interestingly, revealed stronger correlations with encoder-based models, likely due to their bidirectional nature, despite text generation models being primarily based on decoder architectures. Further analysis demonstrated that decoder-based models exhibit more distinct correlations between chosen and rejected responses. Additionally, \acrlong{rms} showed enhanced correlation, when processing both prompts and responses simultaneously.

This research presents the first comprehensive study of an \acrshort{et} corpus for \acrshort{llm} alignment, establishing an important groundwork for future investigations in this domain. Our findings highlight the significance of implicit feedback in user preferences. Additionally, our findings suggest that fine-tuning generative models on this dataset could improve their capacity in human alignment applications. A promising future direction is to enhance human alignment by leveraging implicit feedback from this dataset. We believe the ET data can enhance existing \acrshort{rms} or be used in combination with them.

\subsection{Limitations}

Regarding dataset selection, models utilized - Mistral 7B and Llama - were fine-tuned on publicly available data, though specific details about the training datasets remain limited. While we cannot entirely discard the possibility that OASST1 were included in the models' pretraining, particularly in versions that underwent human alignment, because of the vast scale of training data used in these models, we think that OASST1 inclusion would likely not significantly impact attention correlation. A notable limitation of our study is the relatively small number of participants reading identical texts. Future research would benefit from expanding the participant pool, now that the initial potential has been demonstrated.

%% file: appendix.tex
\appendix

\section*{Appendix}
\section{Additional information on OASST-ETC dataset creation} \label{sec:app:corpus}

\begin{figure}[h]
\centering
 \subfloat{\includegraphics[width=0.7\linewidth]{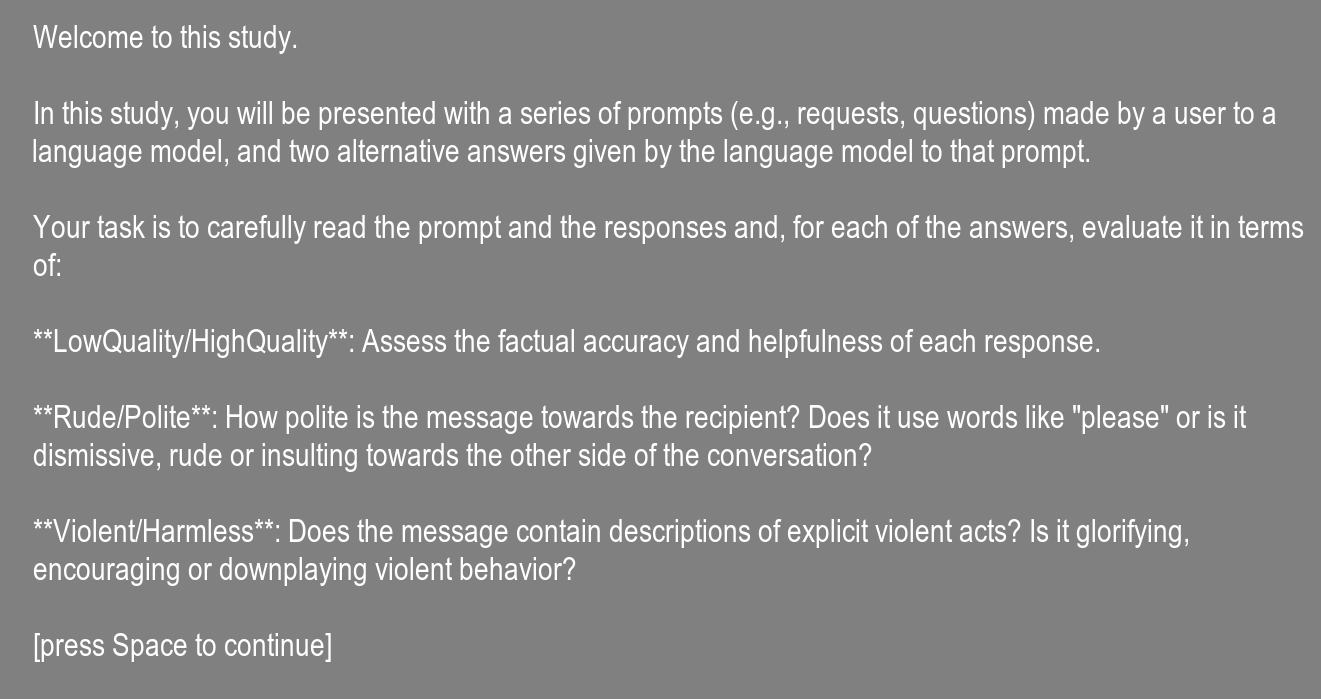}
\label{fig_exp_1}}\\
 \subfloat{\includegraphics[width=0.7\linewidth]{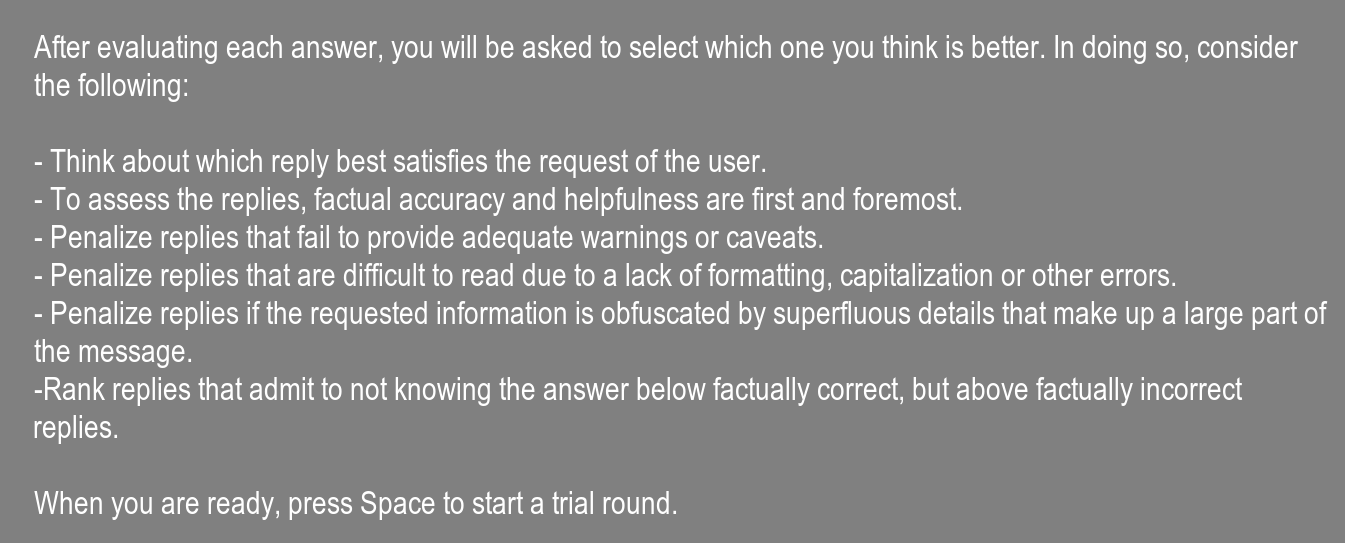}
\label{fig_exp_2}}
\caption{Screenshots of the instructions for the user during the experiment.}
\label{fig:guidelines}
\end{figure}

The \autoref{fig:guidelines} shows screenshots of the instructions that participants read before performing the task.

\begin{figure}[H]
\centering
 \subfloat{\includegraphics[width=0.28\textwidth]{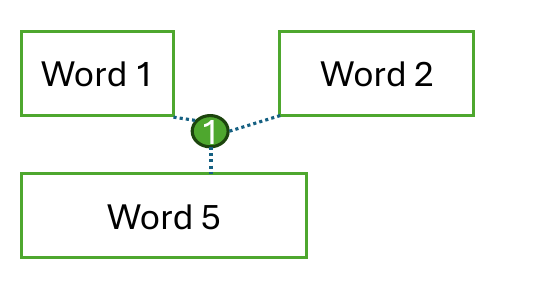}
\label{fig_alg_a}}
 \subfloat{\includegraphics[width=0.55\textwidth]{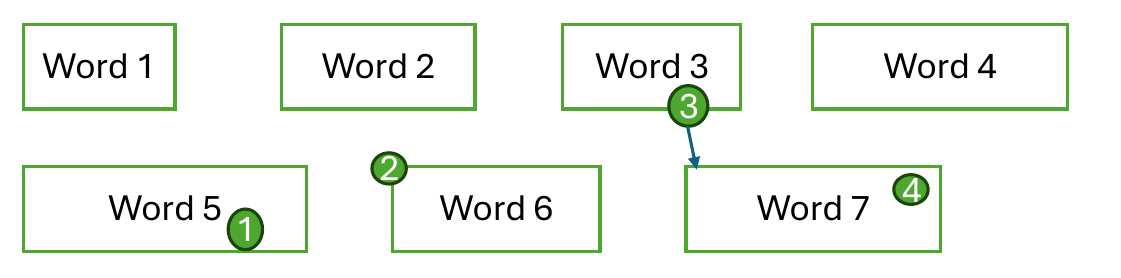}
\label{fig_alg_b}}
\caption{a) Process of assigning a fixation with the X-coordinate within the words b) Process in which a fixation is assigned not to the word with the shortest distance, but to the word on the line where the previous fixation was assigned}
\label{fig:alg_fixations}
\end{figure}

As explained in \autoref{subsec:data_adquisition}, fixations are matched to areas of interest based on Euclidean distance calculations, with the closest area being selected. When calculating distances to words/areas, if a fixation's coordinate falls between two corners, the distance on that axis is zero. Otherwise, distance is measured to the nearest corner. For example, a fixation may have horizontal distance to some words but zero vertical distance to others if it falls between their vertical boundaries. \autoref{fig:alg_fixations} (a) presents an example where the fixation (green dot) has horizontal distance for words 1 and 2 in the X-axis but not for word 5, since it is in the middle of its corners. \autoref{fig:alg_fixations} (b) represented the vertical drift correction where we assign fixations 3 to word 7 instead of word 3, which is the closest one.

\section{Converting token-level features into OCR obtained words-level features} \label{sec:app:mapping_features}

\subsection{Synthetic reading measures} \label{sec:app:mapping_features_syn}
In the generative model, the conversion of word-level features to token-level features during training is done by assigning the features of a word to the first token and it is assumed that the rest of the tokens of this word do not have features. We reversed this process similarly during inference by forcing the predictions for tokens that are not the first in a word to be zero. From each tokenizer we can obtain the list of words from the input text. The way to divide the text into words may depend on the tokenizer. We can also obtain the list of tokens that belong to each word, so we take as reading measures of each word the reading measures of its first token.

For each input text, we have a list of words obtained with the tokenizer on one side, and a list of words obtained by OCR from the screenshot read by the participant on the other side. Generally, these two lists of words should be identical, but in some cases, they might not be. As we explained in \autoref{subsec:eye_features}, each Area of Interest corresponds to a space-separated word, with punctuation marks typically mapped to their adjacent words (\autoref{fig:example_fixations}), so what the tokenizer sees as two words, the OCR might consider as one (and the actual reading measures are considered on this list). Therefore, we map the word lists and in cases where one OCR word corresponds to two tokenizer words, we sum their reading measures. An example of this process can be seen in the \autoref{tab:remapping}.

\begin{table}[h]
    \centering
    \caption{Example of mapping TRT between words obtained by OCR and values obtained by \acrshort{et} generative model.}
    \label{tab:remapping}
    \begin{center}
    \begin{tabular}{c|c|c}
    \toprule
        Words OCR & \multicolumn{2}{c}{unbelievable?}\\\hline
        Words tokenizer & unbelievable & ? \\
        Tokens str & \texttt{['\_un', 'believ', 'able']} & \texttt{['\_?']} \\
        Tokens idx & \texttt{[45-46-47]} & \texttt{[48]} \\
        Tokens IDs & \texttt{[25001-25670-24369]} & \texttt{[25438]} \\
        TRT token-level& \texttt{[10.25, 0, 0]} & \texttt{[2.5]} \\\
        TRT word-level (tokenizer) & \texttt{10.25} & \texttt{2.5} \\\hline
        TRT word-level (OCR) & \multicolumn{2}{c}{12.75} \\

    \end{tabular}
    \end{center}
\end{table}

\subsection{Model-based relative attention} \label{sec:app:mapping_features_att}

From each tokenizer we can obtain the list of words from the input text. The way to divide the text into words may depend on the tokenizer. We can also obtain the list of tokens that belong to each word, so we take as the relative attention of each word the sum of the relative attention of each token associated with the word. Moreover, for each input text, we have a list of words obtained with the tokenizer on one side, and a list of words obtained by OCR from the screenshot read by the participant on the other side. Generally, these two-word lists should be identical, but in some cases they might not be. As we explained in \autoref{subsec:eye_features}, each Area of Interest corresponds to a space-separated word, with punctuation marks typically mapped to their adjacent words (\autoref{fig:example_fixations}), so what the tokenizer sees as two words, the OCR might consider as one. Therefore, we map the word lists and in cases where one OCR word corresponds to two tokenizer words, we sum their relative attention. 

\section{Human- vs. model-based attention measures comparison} \label{sec:app:atenttion}

\subsection{Additional details on the models selected} \label{sec:app:atenttion_models}
\textbf{Transformer encoder-based models:} We use models based on BERT \cite{devlin_bert_2019}. Its bidirectional structure means that each word token is placed in the context of the entire sequence instead of just the tokens appearing before it.

\begin{itemize}
    \item \textbf{bert-base-uncased, bert-base-cased, bert-large-uncased} (BERT) \cite{devlin_bert_2019}: BERT was trained using Masked Language Modeling (MLM), where random words from a sequence are masked during input, and Next Sentence Prediction (NSP), which enables words from one sentence to attend to words in other sentences. We utilized three distinct models: BERT-Base, comprising 12 layers (transformer blocks) and 12 attention heads, available in both cased and uncased variants - the uncased version treats uppercase and lowercase letters identically, such that "dog" and "Dog" are considered equivalent. Additionally, we employed BERT-Large, featuring 24 layers and 16 attention heads, offering enhanced capacity for processing complex patterns and contextual relationships. 
    
    \item \textbf{roberta-base, roberta-large} (RoBERTa) \cite{liu_roberta_2019}: RoBERTa, which maintains BERT's core architecture, implements significant improvements to the training process by eliminating NSP and concentrating solely on masked language modeling with increased pretraining data. Like BERT, RoBERTa is available in both Base and Large configurations.
\end{itemize}

\textbf{Transformer decoder-based models:}
Although most studies focus on bidirectional transformer-encoder based models, we also include decoder-based models to be able to include text generation modes. All these models follow the general structure of decoder-only transformers, which are typically used in autoregressive generation. As explained in \autoref{sec:backgroun_llms}, language model training generally relies on a foundation model trained on massive amounts of data, followed by human alignment processes to align it with users. This process often involves a reward model optimized to evaluate different responses to a prompt. We experiment with both pre-trained models (Mistral 7B and Llama 3) and models that are already human-aligned (Llama 3 Instruct). We also use reward models, explained in \autoref{sec:backgroun_llms}. We chose these models as they are the most used and new open-source models. Among the 3 reward models, all 3 are trained differently based on different backbone models.

\begin{itemize}
    \item \textbf{Phi 1.5} (Phi) \cite{li_textbooks_2023}: The architecture is transformer based with 1.3 billion parameters. It was trained using the same data sources as phi-1, augmented with a new data source that consists of various NLP synthetic texts. When assessed against benchmarks testing common sense, language understanding, and logical reasoning, Phi-1.5 demonstrates a nearly state-of-the-art performance among models with less than 10 billion parameters. The model is not finetuned either for instruction following or through reinforcement learning from human feedback. 
    
    \item \textbf{Llama-2-7b-hf, Llama-2-7b-chat-hf} (Llama-2) \cite{touvron_llama_2023}: Meta released Llama 2 in three sizes: 7B, 13B, and 70B parameters. The architecture uses auto-regressive transformers pretrained on extensive self-supervised data. The 7B parameter version was selected for this work. Llama 2-Chat was developed by applying supervised fine-tuning to the base Llama 2 model, followed by iterative refinement using \acrshort{rlhf} through rejection sampling and \acrshort{ppo}. During \acrshort{rlhf}, they continuously gathered reward modeling data alongside model improvements to maintain in-distribution reward models. 

    \item \textbf{Meta-Llama-3-8B, Llama-3.1-8B, Llama-3.1-8B-Instruct, Meta-Llama-3-8B-Instruct} (Llama 3) \cite{dubey_llama_2024}:  It's a group of language models that naturally handle multiple languages, coding, reasoning, and tool use. Based on Transformer architecture, with the largest version having 405B parameters and up to 128K token context window. In our work, we use the smallest 8B version. These models perform similarly to leading ones like GPT-4 across many tasks and are currently the most powerful open-source LLMs. The post-trained versions build upon the foundation model. Version 3.1's innovation is incorporating image, video, and speech capabilities into Llama 3 through a compositional method. 
    
    \item \textbf{Mistral-7B-v0.1, Mistral-7B-Instruct-v0.1} (Mistral) \cite{jiang_mistral_2023}: Mistral-7B stands as one of the most prominent open-source \acrshort{llm}. Its key innovation lies in the implementation of grouped-query attention (GQA) and sliding window attention (SWA). GQA enhances inference speed and reduces memory consumption during decoding, enabling larger batch sizes and improved throughput - essential features for real-time applications. Additionally, SWA efficiently processes longer sequences while minimizing computational costs, addressing a common constraint in \acrshort{llm}. These combined attention mechanisms enhance Mistral 7B's overall performance and efficiency. For the Instruct version, the model underwent fine-tuning using instruction datasets available through the Hugging Face repository.
    
    \item \textbf{UltraRM-13b} \cite{cui_ultrafeedback_2024}:  UltraRM was trained using LLaMA2-13B \citep{touvron_llama_2023} as the base model. The training combined ULTRAFEEDBACK \cite{cui_ultrafeedback_2024} dataset with other open-source datasets including Stanford SHP, OpenAI Summarization, and Anthropic Helpful \citep{bai_constitutional_2022}. The training strategy and loss objectives remained consistent with \cite{touvron_llama_2023}.
    
    \item \textbf{Eurus-RM-7b} \cite{yuan_advancing_2024}: This model, finetuned from Mistral-7B, is trained on ULTRAINTERACT - a comprehensive dataset designed to enhance LLMs' reasoning abilities. ULTRAINTERACT incorporates diverse instructions from 12 datasets covering mathematics, coding, and logical reasoning, with preference trees for each instruction. Eurus-RM-7B combines ULTRAINTERACT with UltraFeedback and UltraSafety datasets, using a specialized reward modeling objective. The model's training approach is designed to improve reasoning capabilities, especially for complex problems, while maintaining balance across different reward modeling abilities through dataset integration. The original model architecture from the Huggingface Checkpoint is used with minimal modifications to extract layer attention during inference. Additional details will be available in the open repository upon paper publication.
    
    \item \textbf{QRM-Llama3.1-8B} \cite{dorka_quantile_2024}: Is based on the Llama3.1-8B model foundation. The backbone architecture stays fixed during training, allowing efficient computation through single-pass feature extraction and reuse. The multi-attribute regression incorporates 19 distinct attributes from eight datasets. Training utilizes Quantile Reward Models (QRMs), an innovative approach learning complete reward distributions instead of single values. Through quantile regression, QRMs capture the full range of preference distributions, including multiple modes, providing a deeper understanding of human preferences. This method handles preference diversity, label noise, and contradictory preferences by representing them as distinct distribution modes. The original model architecture from the Huggingface Checkpoint is used with minimal modifications to extract layer attention during inference. Additional details will be available in the open repository upon paper publication.
\end{itemize}
     
\subsection{Implementation details on the correlation analysis of \acrshort{rm} input} \label{sec_app_reward_model}

As explained in \autoref{sec:backgroun_llms}, the general input of the reward model consists of combining a prompt with a response, where the output is a reward measuring how good that response is for the prompt. The concatenation of the prompt and response is typically done using special tokens, which usually depend on how the model was trained. The special tokens differentiate the turns between the user and the system.  For each of the three models, we have used the tokens proposed in their implementations, which we show below. Additional implementation details can be found in the open-source code.

\begin{itemize}

    \item \textbf{QRM-Llama3.1-8B:}
    \textit{ \textless start\_header\_id\textgreater user\textless end\_header\_id\textgreater} \textcolor{blue}{Example Prompt} \textit{\textless eot\_id\textgreater \textless start\_header\_id\textgreater assistant \textless end\_header\_id\textgreater} \textcolor{blue}{Example Response} 

    \item \textbf{UltraRM-13b:} \textit{Human:} \textcolor{blue}{Example Prompt} \textit{Assistant:} \textcolor{blue}{Example Response}

    \item \textbf{Eurus-RM-7b:} \textit{ [INST]} \textcolor{blue}{Example Prompt} \textit{[/INST]} \textcolor{blue}{Example Response} 
\end{itemize}

\subsection{Additional results on correlation analysis of model layers and reading measures} 
\label{app:sec:analuse_layers}
\begin{figure}[h!]
    \centering
    \subfloat{
        \includegraphics[width=0.20\textwidth]{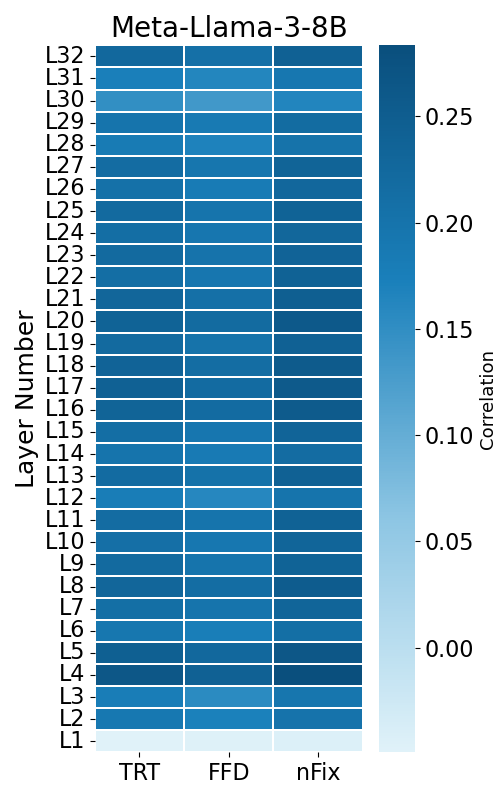}
        \label{fig_alg_a1}
    }
  \subfloat{
        \includegraphics[width=0.20\textwidth]{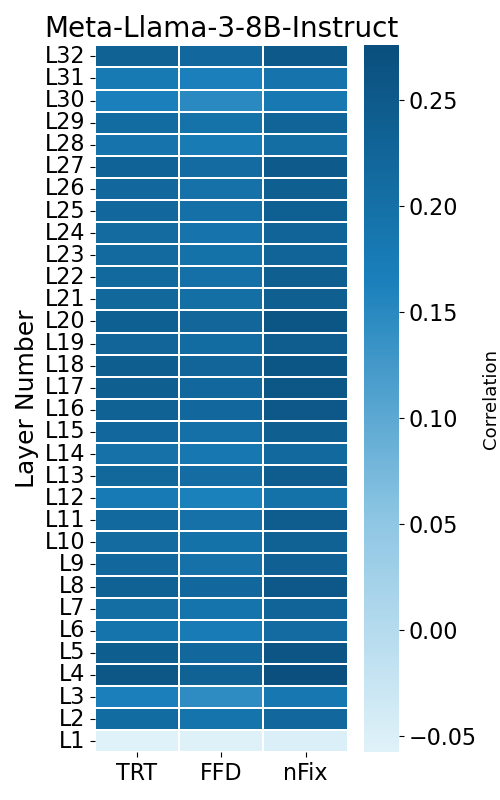}
        \label{fig_alg_b1}
    }
  \subfloat{
        \includegraphics[width=0.20\textwidth]{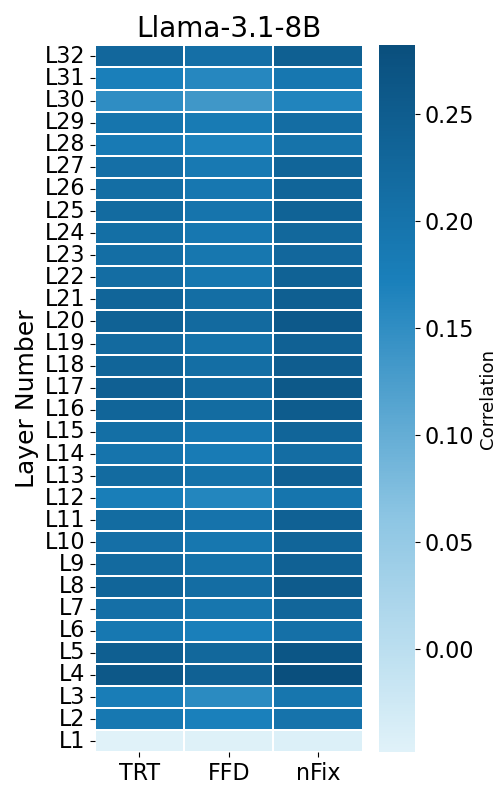}
        \label{fig_alg_c1}
    }
    \subfloat{
        \includegraphics[width=0.20\textwidth]{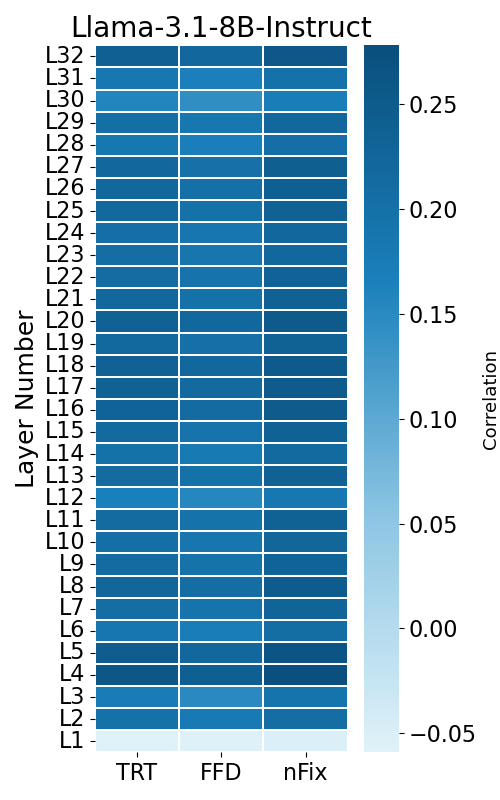}
        \label{fig_alg_d1}
    }
    
    \subfloat{
        \includegraphics[width=0.20\textwidth]{figures/att_layer/Llama-2-7b-hf_attention_layers_all_trials.png}
        \label{fig_alg_a2}
    }
    \subfloat{
        \includegraphics[width=0.20\textwidth]{figures/att_layer/Llama-2-7b-chat-hf_attention_layers_all_trials.png}
        \label{fig_alg_b2}
    }
    \subfloat{
        \includegraphics[width=0.20\textwidth]{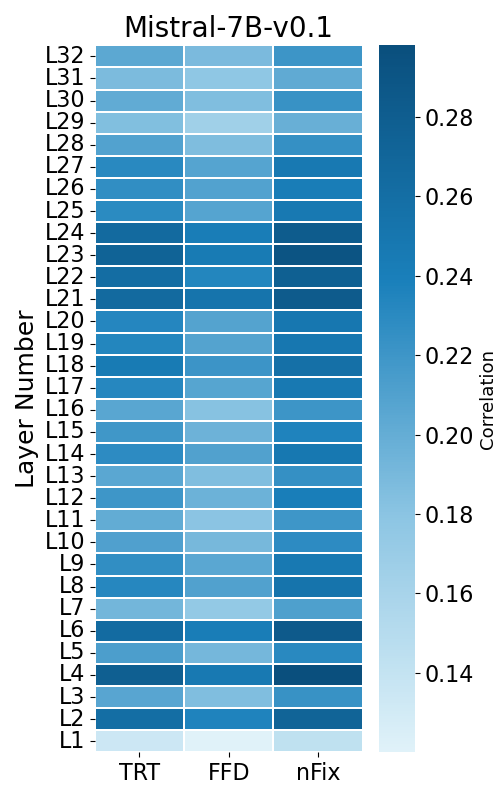}
        \label{fig_alg_c2}
    }
    \subfloat{
        \includegraphics[width=0.20\textwidth]{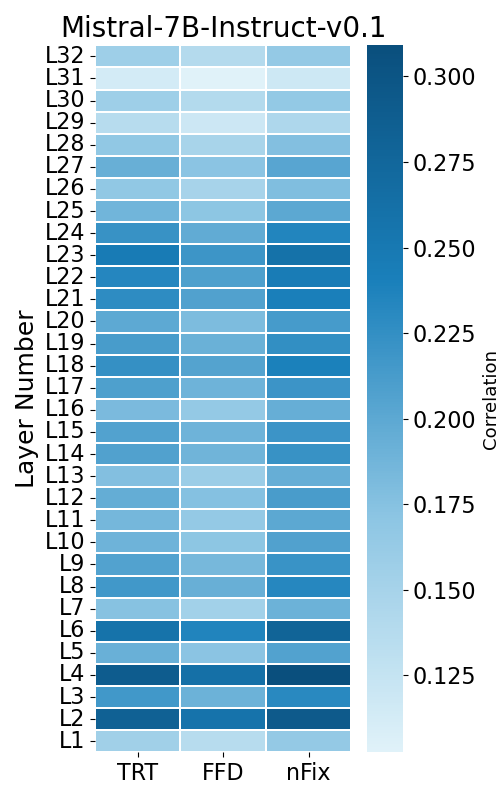}
        \label{fig_alg_d2}
    }
    
    \subfloat{
        \includegraphics[width=0.20\textwidth]{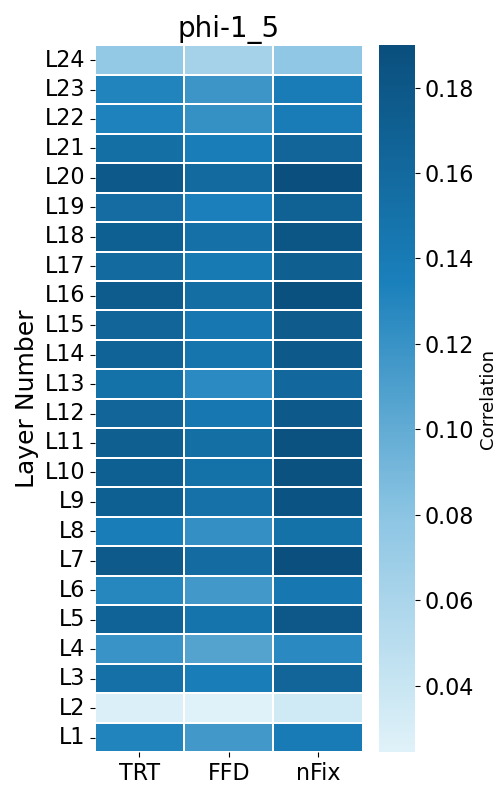}
        \label{fig_alg_a3}
    }
    \subfloat{
        \includegraphics[width=0.20\textwidth]{figures/att_layer/Eurus-RM-7b_attention_layers_all_trials.png}
        \label{fig_alg_b3}
    }
    \subfloat{
        \includegraphics[width=0.20\textwidth]{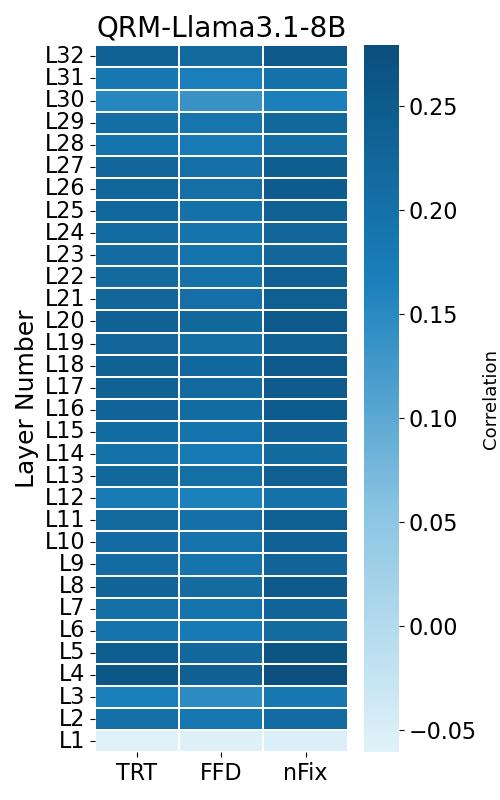}
        \label{fig_alg_c3}
    }
    \subfloat{
        \includegraphics[width=0.20\textwidth]{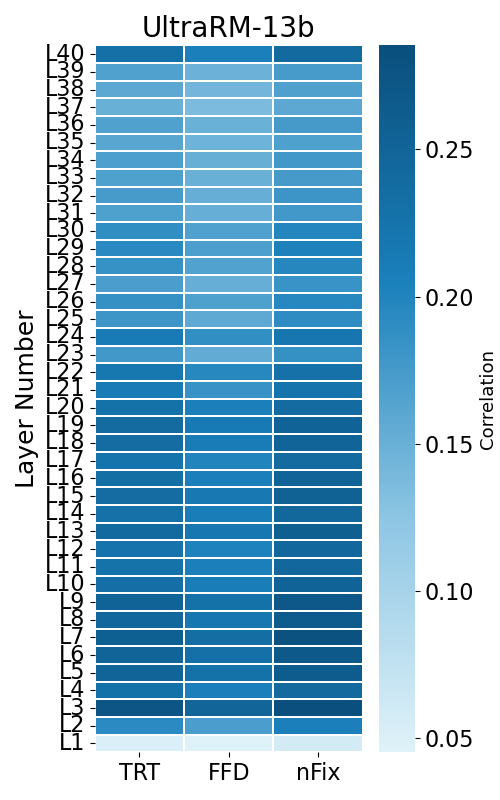}
        \label{fig_alg_d3}
     }   
     
    \subfloat{
        \includegraphics[width=0.18\textwidth]{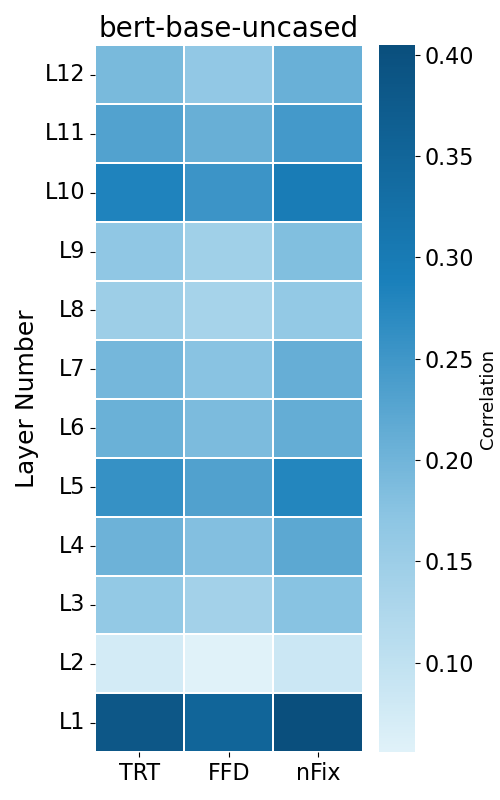}
        \label{fig_alg_a3}
    }
    \subfloat{
        \includegraphics[width=0.18\textwidth]{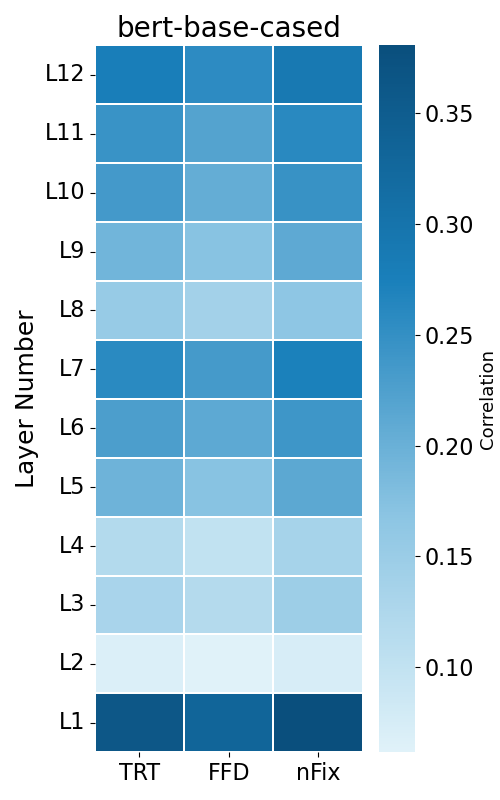}
        \label{fig_alg_b3}
    }
    \subfloat{
        \includegraphics[width=0.18\textwidth]{figures/att_layer/bert-large-uncased_attention_layers_all_trials.png}
        \label{fig_alg_c3}
    }
    \subfloat{
        \includegraphics[width=0.18\textwidth]{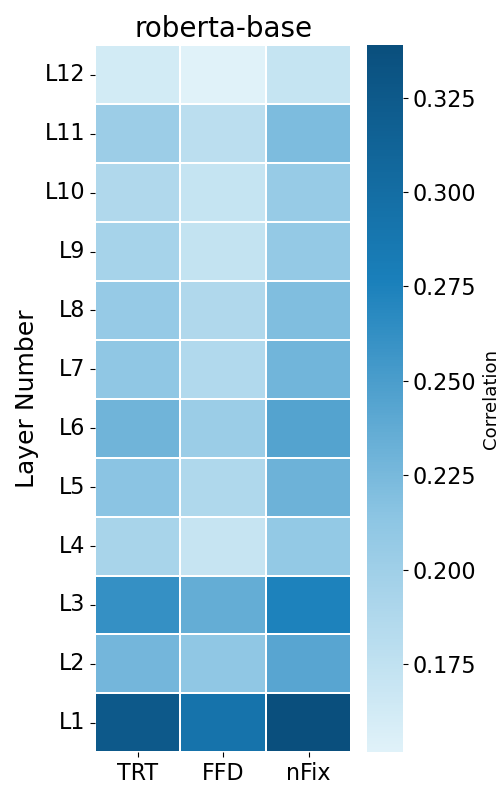}
        \label{fig_alg_d3}
    }
    \subfloat{
        \includegraphics[width=0.18\textwidth]{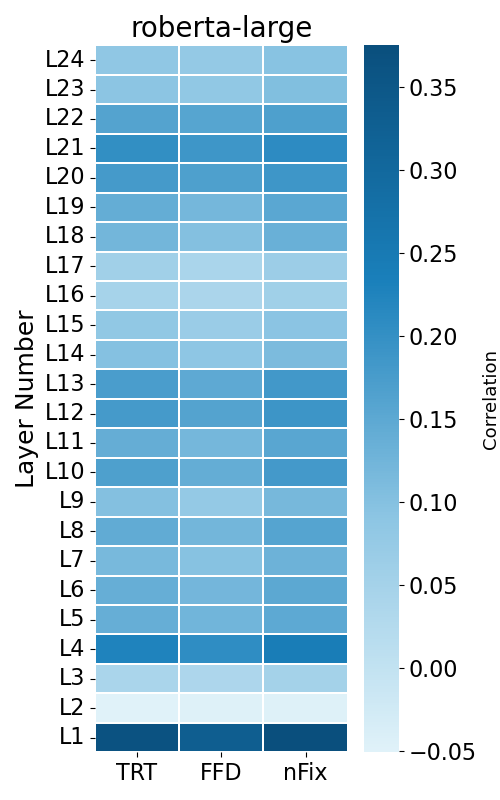}
        \label{fig_alg_d3}
    }   
    
    \caption{Mean Spearman correlation values of different layers in different models with TRT, FFD and nFix reading measures for all models.}
    \label{fig:att_layer_all}
\end{figure}

In \autoref{fig:att_layer_all}, the results complementary to \autoref{fig:att_layer}, in \autoref{sec:analuse_layers} can be seen.

\subsection{Additional results on the correlation analysis of model architectures and reading measures} 
\label{app:sec:analyse_models}

\begin{figure}[h!]
    \centering
     \subfloat{\includegraphics[width=0.48\textwidth]{figures/att_model/trials_TRT_n_f_barplot_means_stds_chosenrejected.png}
    \label{fig_att_model_TRT_c}}
     \subfloat{\includegraphics[width=0.48\textwidth]{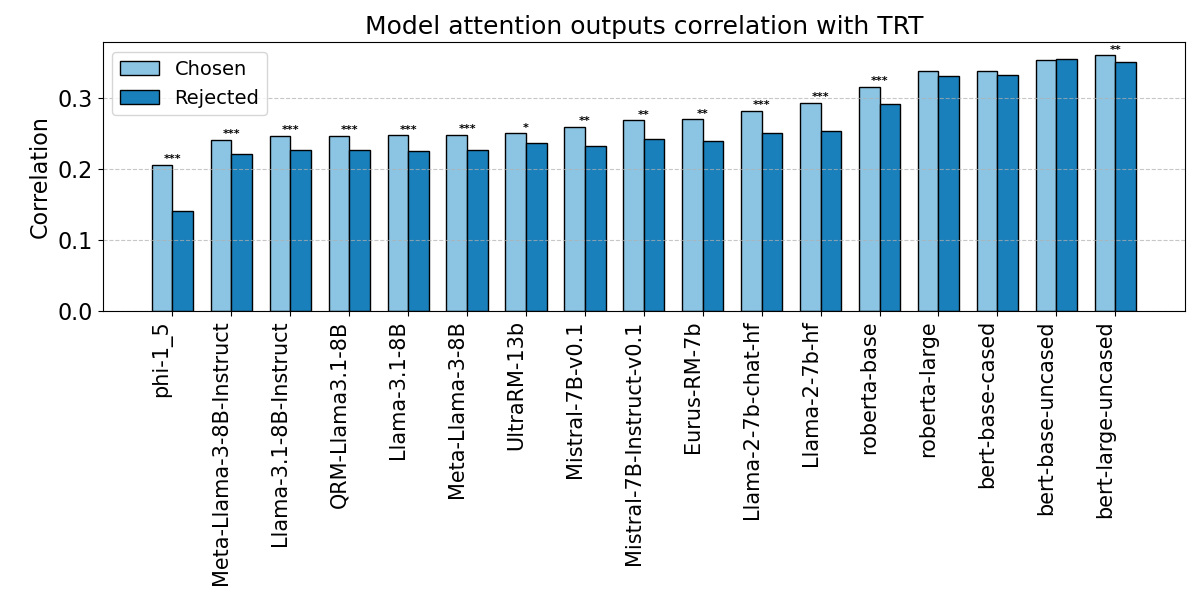}}
    \label{fig_att_model_TRT_all}
    \caption{Mean Spearman correlation analysis between TRT and different models. Left: Only trials read by three participants are included. Right: All trials are included. * indicates statistical significance between chosen and rejected.}
    \label{fig_att_model_TRT_appendix}
\end{figure}

\begin{figure}[h!]
    \centering
     \subfloat{\includegraphics[width=0.48\textwidth]{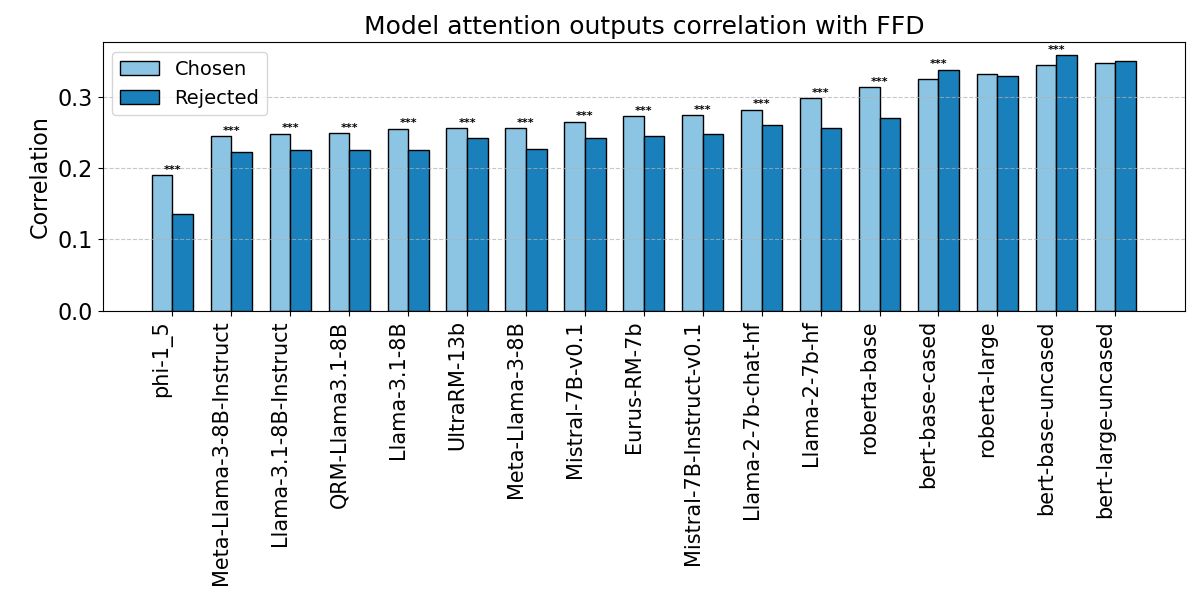}
    \label{fig_att_model_FFD_c}}
     \subfloat{\includegraphics[width=0.48\textwidth]{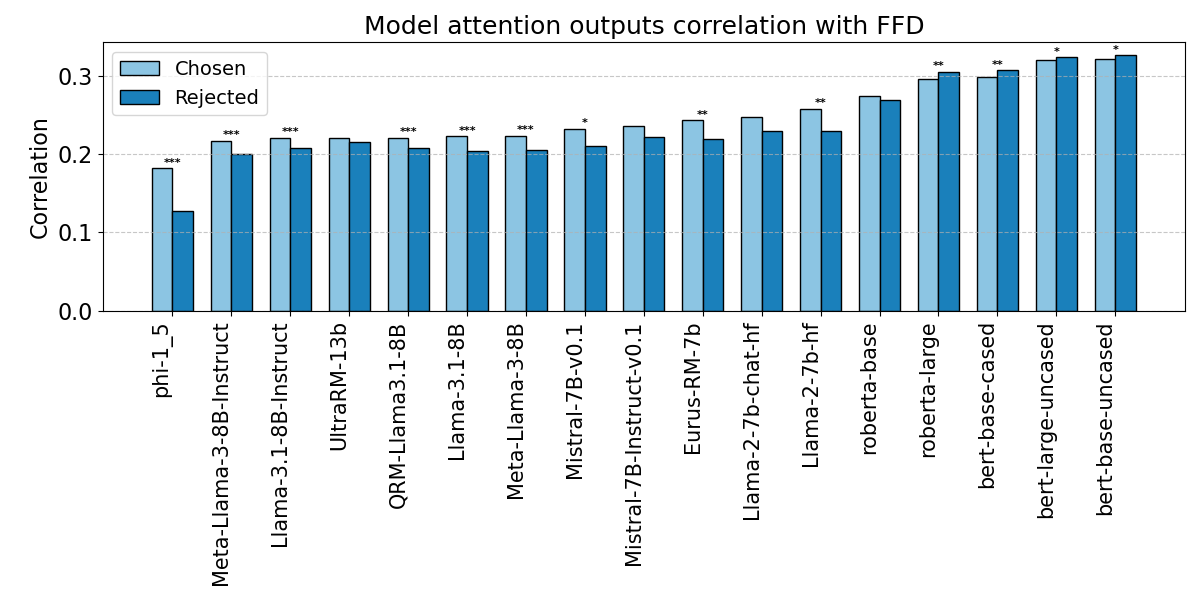}}
    \label{fig_att_model_FFD_all}
    \caption{Mean Spearman correlation analysis between FFD and different models. Left: Only trials read by three participants are included. Right: All trials are included. * indicates statistical significance between chosen and rejected.}
    \label{fig_att_model_FFD_appendix}
\end{figure}

\begin{figure}[h!]
    \centering
     \subfloat{\includegraphics[width=0.48\textwidth]{figures/att_model/trials_nFix_f_barplot_means_stds_chosenrejected.png}
    \label{fig_att_model_nfix_c}}
     \subfloat{\includegraphics[width=0.48\textwidth]{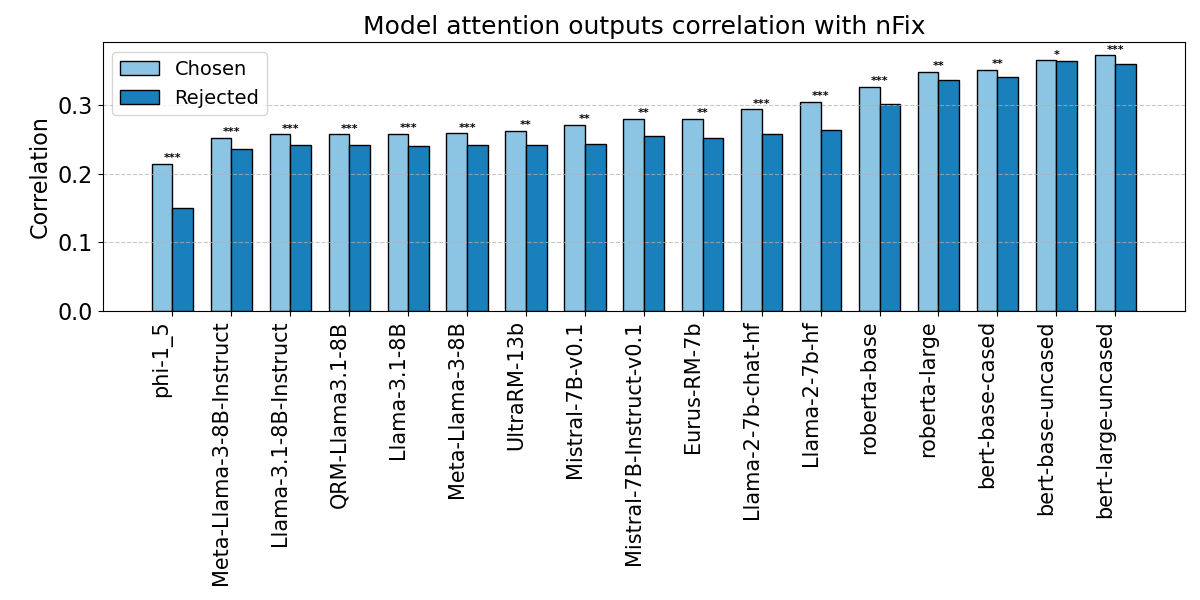}}
    \label{fig_att_model_nfix_all}
    \caption{Mean Spearman correlation analysis between nFix and different models. Left: Only trials read by three participants are included. Right: All trials are included. * indicates statistical significance between chosen and rejected.}
    \label{fig_att_model_nfix_appendix}
\end{figure}

In \autoref{fig_att_model_TRT_appendix}, \autoref{fig_att_model_FFD_appendix} and \autoref{fig_att_model_nfix_appendix} complementary results are shown to the ones in \autoref{sec:analyse_models}.

\subsection{Additional results on the correlation analysis of \acrshort{rm} input and reading measures} \label{app:sec:reward_modelling} 

\begin{figure}[h!]
    \centering
    \subfloat{\includegraphics[width=0.30\textwidth]{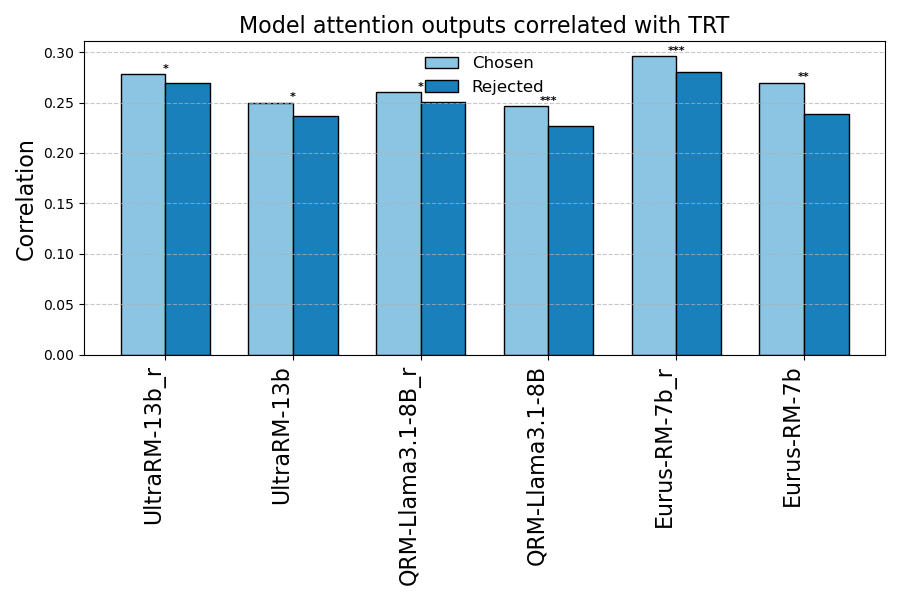}}
    \label{fig_att_model_trt_reward}
     \subfloat{\includegraphics[width=0.30\textwidth]{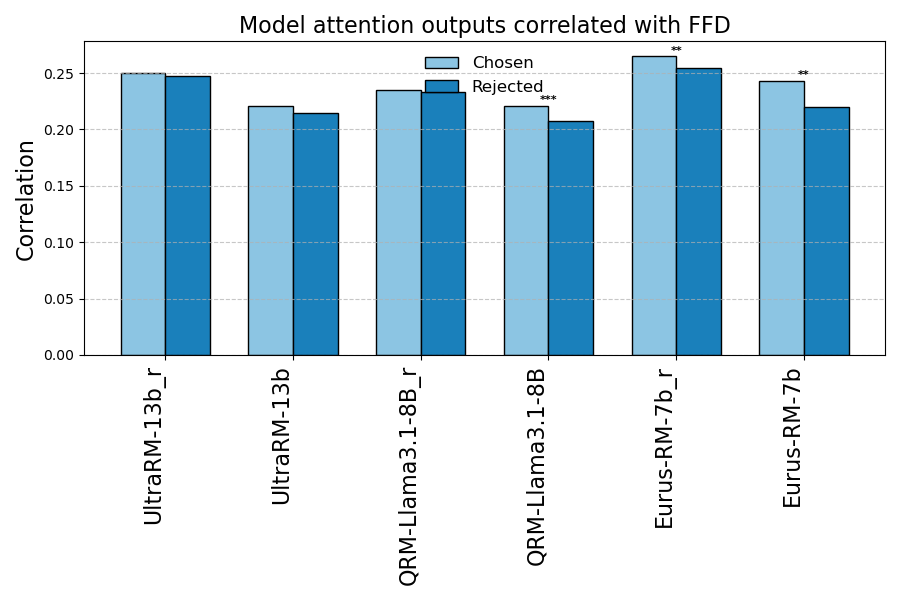}}
    \label{fig_att_model_ffd_reward}
    \subfloat{\includegraphics[width=0.30\textwidth]{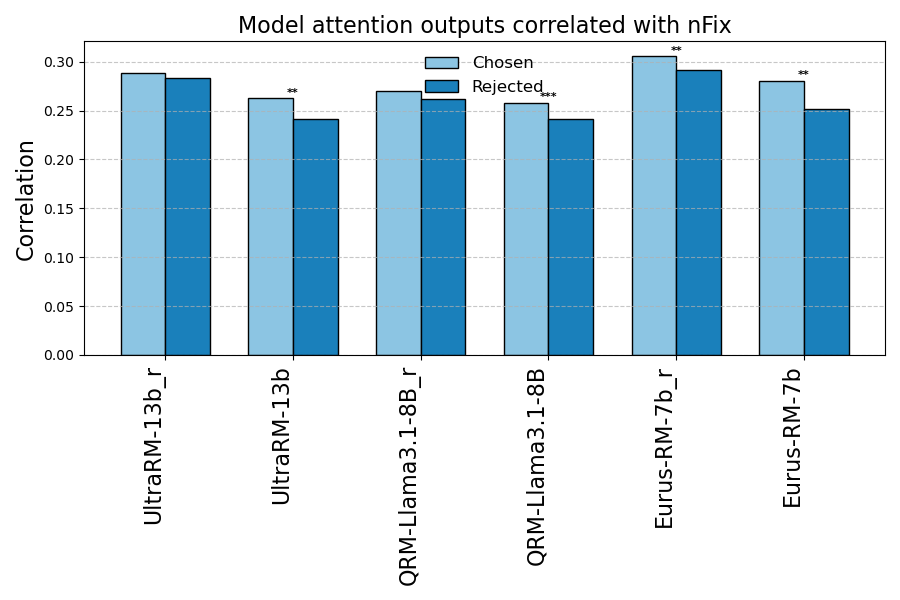}}
    \caption{Mean Spearman correlation coefficients comparing TRT, FFD, and nFix across three reward models for all trials, with both standard and prompt-conditioned (\_r) variations. * indicates statistical significance between chosen and rejected.}
    \label{fig_att_model_reward_appendix}
\end{figure}

In \autoref{fig_att_model_reward_appendix} we show similar results to \autoref{fig_att_model_reward} in \autoref{sec:reward_modelling} but for all trials, not only the unanimous responded ones.

\subsection{Hardware}
\label{sec:hadware}

 We use servers equipped with 2x Intel Xeon Platinum 8470 CPUs, 1TB of RAM and 2X NVIDIA A100 (80GB) GPUs. At inference time we were using only one GPU.

\section{EDA acquisition} 
\label{sec:app_eda}

The electrodermal activity (EDA) was measured using the Versatile Bio acquisition system (Bitbrain, Zaragoza, Spain). Previous literature explained the inclusion of EDA as a reliable metric, stating it is one of the most reliable and usable psychophysiological parameters for monitoring the cognitive state \cite{bontchev2016assessing}. For the current study, two electrodes were placed on the tip of the index and middle finger of the participant's non-dominant hand. The EDA was collected by using a pair of Ag/AgCl electrodes, which provide a reliable and stable skin conductance measurement due to the fact that sweat gland activation (indicating mental state fluctuations) creates variations in the signal. The skin conductance then is expressed in microsiemens (default measurement unit), which shows the facility with which the skin allows electrical current to pass through it.

The signal was sampled at a rate of 256 Hz, ensuring the possibility to capture both tonic and phasic components. For this project, as we concentrated in event-related responses, we focused on phasic activity (also known as skin conductance responses) which occur as a response to specific stimuli. For this goal, a baseline is subtracted to normalise the offset of the analysed data. 

Synchronising the data streams from different sources, such as EDA from OpenVibe and behavioural markers from PsychoPy, introduces certain challenges. For instance, timing discrepancies between systems can lead to misaligned data, which we addressed by using precise onset markers embedded directly in the data stream. This ensured accurate synchronisation across modalities. Furthermore, we combined several devices working with different data formats, which we processed and aligned to have a complete neurophysiological and behavioural profile of each individual response.
\begin{table}
\caption{Group-Level Analysis of Preferred and Non-Preferred EDA Conditions}
\label{tab:eda_group_analysis}
\begin{tabular}{|l|c|c|}
\toprule
 & Chosen EDA (Mean ± SD) & Rejected EDA (Mean ± SD) \\
\midrule
\textbf{Mean} & 3.06 ± 1.98 & 3.05 ± 1.97 \\
\textbf{Standard Deviation} & 0.09 ± 0.05 & 0.08 ± 0.04 \\
\textbf{Minimum} & 2.86 ± 1.91 & 2.87 ± 1.90 \\
\textbf{Maximum} & 3.22 ± 2.05 & 3.20 ± 2.04 \\
\textbf{Median} & 3.06 ± 1.99 & 3.06 ± 1.97 \\
\textbf{Range} & 0.37 ± 0.21 & 0.33 ± 0.20 \\
\bottomrule
\end{tabular}
\end{table}

\section{Notes on statistical analysis} 
\label{app:stats_notes}

In \autoref{fig:words_lenght}, \autoref{fig:reading_measure_real}, \autoref{fig:reading_measure_synthe}, \autoref{fig_att_model}, and \autoref{fig_att_model_reward} represent statistical significance levels of $ * p < 0.05 $,  $ ** p < 0.01 $ and $ *** p < 0.001 $. Regarding the metrics reported in \autoref{sec:compute_readingm} and in \autoref{fig:reading_measure_real} and \autoref{fig:reading_measure_synthe}, as highlighted in \autoref{tab:data_samples}, the total number of annotated responses is 652. Therefore, when we compare the preferred versus the rejected (we divide by two since we have two conditions), we have 652/2 - 1 = 325. In the case of synthetic measures, we have 321, as we could not generate synthetic reading measures for some responses due to typos. For the unanimously annotated responses, the same logic was used: 214/2 - 1 = 106. 

\section{Notes on dataset annotation}
\label{app:annotations}

The original dataset (OASST1) was annotated by crowdsourced volunteers who carried out multiple tasks such as ranking assistant replies, labeling messages for quality and safety, and providing preference ratings. Each message was evaluated by at least three annotators, and they used Tideman’s Ranked-Pairs method to consolidate these rankings. The annotations comprised both Likert-scale ratings (covering aspects like quality, creativity, and politeness) and binary labels (identifying factors like spam, hate speech, and guideline violations). This rigorous process helped ensure high-quality, human-driven annotations that have been central to both training and evaluating language models in the literature. Moreover, we took an additional step by selecting the most distant responses in the ranking for each prompt, thereby maximizing the contrast in participant preferences and we acknowledge that less distant responses may not yield the same clear effects. It is important to note that annotating responses naturally involves some degree of subjectivity. To address this, we conducted a similar procedure with our own preference judgments and compared them to the original labels, reporting the results ( \autoref{tab:agreement}) as the proportion of trials (between 0 and 1) in which both labels matched.

\begin{table}[h!]
    \centering
    \caption{Agreement between participants and original dataset labels.}
    \label{tab:agreement}
    \begin{tabular}{c|ccccccccc}
    \toprule
    & \textbf{total} & \textbf{1} & \textbf{2} & \textbf{3} & \textbf{4} & \textbf{5} & \textbf{6} & \textbf{7} & \textbf{8} \\ 
    \hline
    \textbf{all} & 0.58 & 0.70 & 0.68 & 0.42 & 0.56 & 0.58 & 0.50 & 0.66 & 0.57 \\
\textbf{unanimous annotated}  &0.59 & 0.62 & 0.76 & 0.44 & 0.57 & 0.60 & 0.50 & 0.64 & 0.60 \\
    \bottomrule
    \end{tabular}
\end{table}